%% file: fdnlp_paper.tex
\documentclass[11pt,twocolumn,logo,copyright]{fdnlp_main}

\usepackage{microtype}
\usepackage{xurl}
\usepackage{graphicx} 
\usepackage{float}
\usepackage{natbib}  
\usepackage{caption} 
\usepackage{algorithm}
\usepackage{algorithmic}

\usepackage{newfloat}
\usepackage{amssymb}
\usepackage{listings}
\DeclareCaptionStyle{ruled}{labelfont=normalfont,labelsep=colon,strut=off} 
\floatstyle{ruled}
\newfloat{listing}{tb}{lst}{}
\floatname{listing}{Listing}

\usepackage{booktabs}
\usepackage{placeins}
\usepackage{siunitx}
\usepackage{pifont}
\usepackage[most]{tcolorbox}

\renewcommand{\topfraction}{0.95}
\renewcommand{\dbltopfraction}{0.95}
\renewcommand{\textfraction}{0.05}
\renewcommand{\floatpagefraction}{0.80}
\renewcommand{\dblfloatpagefraction}{0.80}

\definecolor{gblue9}{RGB}{23,78,166}

\providecommand{\shortcite}[1]{\citep{#1}}

\title{\textsc{Sci-MMR}: Benchmarking Multi-Step Evidence-Grounded Scientific Reasoning in Multimodal Agents}
\author{
    Jiaqiang Li\textsuperscript{*},
    Yajie Yang\textsuperscript{*},
    Zhiheng Xi\textsuperscript{*},
    Jiadong Chen,
    Enyu Zhou,
    Senjie Jin,
    Yang Nan,
    Jiazheng Zhang,
    Han Wang,
    Yanxin Li,
    Dingwei Zhu,
    Bicheng Deng,
    Yuhui Wang,
    Xiang Zheng,
    Qi Zhang,
    Lei Bai,
    Xingjun Ma\textsuperscript{\textdagger},
    Tao Gui\textsuperscript{\textdagger}\\
    \vspace{0.3cm}
    \normalsize Fudan NLP Group\\
}

\begin{document}

\begin{abstract}
Autonomous research agents are increasingly expected to search the literature, analyze experimental evidence, and generate scientific hypotheses. These capabilities require multi-step evidence-grounded reasoning that progressively acquires, integrates, and verifies evidence before reaching a conclusion. Existing multimodal benchmarks, however, largely evaluate final-answer accuracy, leaving open whether predictions are actually supported by traceable scientific evidence. We introduce \textbf{\textsc{Sci-MMR}}, a benchmark for multi-step evidence-grounded scientific reasoning built on structured argument graphs linking scientific claims, citation-grounded knowledge, visual evidence, and supporting regions. Sci-MMR comprises 235 multi-hop reasoning tasks spanning four scientific disciplines, with an average of nine figure panels per task. Evaluating eight frontier multimodal models, we find that answer accuracy consistently exceeds complete-evidence recovery rate by more than 20\%, revealing a substantial gap that answer-only evaluation is structurally unable to capture. Through controlled interventions, we identify two fundamental bottlenecks. First, evidence acquisition: models struggle to extract complete structured evidence from scientific figures, accounting for 57.2\% of failures. While cropping tools yield modest gains (+4.5 points), providing gold evidence improves accuracy by up to 37.0 points, indicating difficulty in assembling complete multi-region evidence. Second, evidence integration: models struggle to translate available evidence into correct conclusions, accounting for 31.8\% of failures, while even with gold evidence the strongest model achieves only 69.1\% accuracy on the hardest tasks. These findings indicate that current answer-centric benchmarks substantially overestimate the evidence-grounded reasoning capabilities of multimodal research agents.

\end{abstract}

\maketitle
\fdnlpauthornotes


\section{Introduction}
Scientific reasoning rarely relies on isolated facts. Instead, researchers progressively acquire, integrate, and verify evidence from textual descriptions, visual observations, and experimental results before reaching a scientific conclusion~\citep{DBLP:conf/nips/PramanickCV24,DBLP:conf/emnlp/LiS0L0C24}. As autonomous research agents become increasingly capable of literature analysis, experimental interpretation, and hypothesis generation~\citep{DBLP:journals/corr/abs-2408-06292,
DBLP:conf/emnlp/SchmidgallSWSWYLMLB25,
DBLP:journals/corr/abs-2502-18864}, multi-step evidence-grounded reasoning becomes a fundamental capability. Unlike conventional multimodal question answering, these agents must actively acquire evidence distributed across figures, experiments, and prior scientific knowledge before synthesizing reliable conclusions. Existing multimodal evaluations, however, primarily measure final-answer correctness~\citep{DBLP:conf/nips/PramanickCV24,
DBLP:conf/acl/WangSKC025,
DBLP:conf/acl/ZhaoWZNBCRYTH26}, leaving unanswered whether models can actually acquire and organize the evidence needed to justify their predictions.

\begin{figure}[t]
      \centering
      \includegraphics[width=\columnwidth]{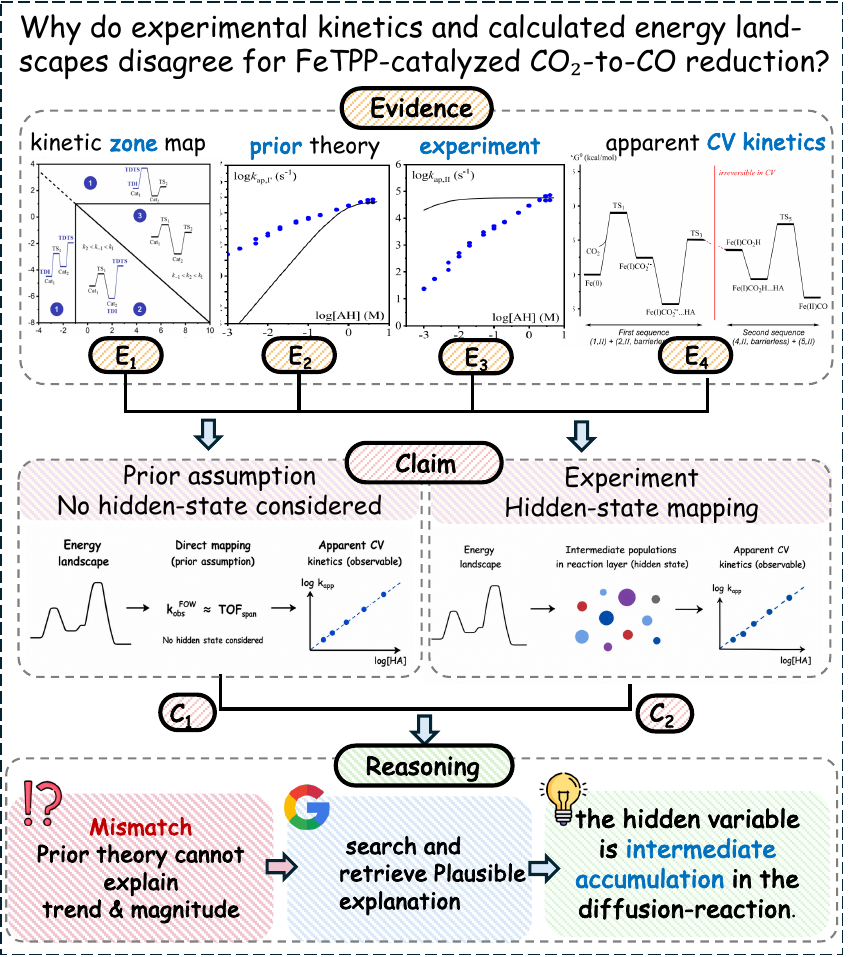}
       \caption{\textbf{Illustration of a \textsc{Sci-MMR} task.} A scientific reasoning task requiring evidence acquisition across multiple figure panels, intermediate-claim reasoning, knowledge retrieval, and premise auditing to reconcile experimental kinetics with calculated energy landscapes.}
      \label{fig:teaser}
\end{figure}

To address this gap, we introduce \textsc{Sci-MMR}, a benchmark for evaluating multi-step evidence-grounded scientific reasoning in multimodal agents. Built from peer-reviewed scientific publications, \textsc{Sci-MMR} represents each task as a structured argument graph linking scientific claims, citation-grounded knowledge, visual evidence, and supporting image regions, enabling fine-grained evaluation of both evidence acquisition and evidence integration (Figure~\ref{fig:teaser}). The benchmark comprises 235 multi-hop reasoning tasks spanning four scientific disciplines and 35 domains, with each task requiring evidence aggregation across an average of nine figure panels rather than a single figure or table.

Evaluating eight frontier multimodal models under four reasoning settings---\textit{Caption-only}, \textit{Direct Visual Reasoning}, \textit{Evidence-Hint Reasoning}, and \textit{Agentic Tool-Use}---reveals a consistent gap between answer accuracy and evidence coverage: models frequently reach correct conclusions while recovering only partial supporting evidence. Through controlled interventions, we identify two fundamental bottlenecks. 
The first is evidence acquisition. Models often identify relevant visual regions but fail to assemble the complete evidence distributed across multiple figures. Under Direct Visual Reasoning, evidence access and grounding account for 57.2\% of errors. Models locate at least one relevant region in 62.1\% of runs, but recover the complete evidence set in only 12.7\%, showing that the challenge lies not in finding any relevant region but in assembling the complete set that multi-panel tasks require. Equipping models with visual cropping tools raises accuracy by 4.5 points, confirming that active localization does help, though its gains remain modest compared to gains of 19.1--37.0\% from directly supplying gold evidence statements -- indicating that most of the remaining difficulty lies in achieving complete, multi-region localization, not in extracting evidence once a region is found.

The second bottleneck is evidence integration. models struggle to translate available evidence into correct conclusions, accounting for 31.8\% of failures, while even with gold evidence the strongest model achieves only 69.1\% accuracy on the hardest tasks. These findings show that evidence-grounded reasoning remains a distinct challenge beyond evidence acquisition, even when the required evidence is fully available.

Our main contributions are summarized as follows:
\begin{itemize}
    \item We introduce \textsc{Sci-MMR}, the first benchmark for multi-step evidence-grounded scientific reasoning. Built from peer-reviewed publications, it comprises 235 multi-hop reasoning tasks spanning four scientific disciplines and 35 domains, grounded in structured argument graphs connecting scientific claims, citation-grounded knowledge, visual evidence, and supporting image regions.
    
    \item We develop a four-setting evaluation protocol that disentangles failures in evidence acquisition, evidence integration, and final-answer reasoning, enabling fine-grained diagnosis of multimodal scientific reasoning.
    
    \item Our analysis of 8 frontier multimodal models reveals two fundamental bottlenecks: 1) they struggle to transform complex scientific figures into complete structured evidence; and 2) evidence integration remains challenging even when gold evidence is provided, indicating that evidence-grounded reasoning extends well beyond visual perception.
\end{itemize}

\newcommand{\cmark}{\ding{51}}
\newcommand{\xmark}{\ding{55}}

\section{Related Work}
\label{sec:related-work}

\paragraph{Synthesizing Multi-Hop Reasoning Tasks.}
Existing scientific multimodal benchmarks evaluate figure-grounded question answering, claim verification, and paper-level understanding, but typically treat reasoning as recovering a final answer rather than reconstructing the evidence supporting a scientific claim~\citep{DBLP:conf/nips/PramanickCV24,DBLP:conf/acl/WangSKC025,DBLP:journals/corr/abs-2604-01306,DBLP:conf/acl/ZhaoWZNBCRYTH26}. Existing multi-hop task construction methods compose reasoning steps within a single image~\citep{DBLP:journals/corr/abs-2603-17024,DBLP:conf/iccv/TranTHP25}, sample paths from knowledge or content graphs~\citep{DBLP:journals/corr/abs-2603-00873,DBLP:journals/corr/abs-2604-12890,DBLP:journals/corr/abs-2604-01634}, or synthesize retrieval trajectories through graph expansion and information obfuscation~\citep{DBLP:journals/corr/abs-2507-02592}. In contrast, \textsc{Sci-MMR} reconstructs evidence dependencies directly from peer-reviewed papers, grounding each reasoning step in citation-supported evidence and visual regions.

\paragraph{Beyond Answer-Centric Evaluation.}
Recent work has moved beyond answer-only evaluation by assessing intermediate reasoning and agent behaviors. PhysicsArena~\citep{DBLP:conf/emnlp/DaiYSZGHLZTGH25} evaluates variable identification, process formulation, and solution derivation, while VDR-Bench~\citep{DBLP:journals/corr/abs-2602-02185} measures intermediate entity recovery alongside answer correctness. Research-agent benchmarks further introduce expert-authored rubrics, process--report consistency, capability-aware evaluators, task-specific judging, and multimodal evidence-fidelity checks~\citep{DBLP:journals/corr/abs-2511-07685,DBLP:journals/corr/abs-2603-28407,DBLP:conf/acl/Ben-AvrahamLDGN26,DBLP:journals/corr/abs-2603-29139,DBLP:journals/corr/abs-2601-12346}. These approaches reveal intermediate failures beyond final-answer accuracy but do not evaluate whether models acquire the complete evidence supporting a scientific conclusion. \textsc{Sci-MMR} instead disentangles evidence acquisition from evidence integration by evaluating evidence coverage and evidence-to-claim reasoning as complementary signals.

\begin{table}[t]
\centering
\setlength{\tabcolsep}{2.3pt}

\resizebox{\columnwidth}{!}{%
\begin{tabular}{
  l
  c
  S[table-format=2.1]
  S[table-format=1.1]
  c
  c
}
\toprule
\textbf{Benchmark}
& \textbf{Disc.}
& \multicolumn{1}{c}{\textbf{Img.}}
& \multicolumn{1}{c}{\textbf{Hop}}
& \textbf{Agentic}
& \textbf{Process} \\
\midrule
SPIQA~\shortcite{DBLP:conf/nips/PramanickCV24}
& 1 & 10.3 & 1.0 & \xmark & \xmark \\

SciVer~\shortcite{DBLP:conf/acl/WangSKC025}
& 1 & 1.5 & 1.0 & \xmark & \xmark \\

PhysicsArena~\shortcite{DBLP:conf/emnlp/DaiYSZGHLZTGH25}
& 1 & 1.0 & 1.0 & \xmark & \cmark \\

PaperMind~\shortcite{DBLP:conf/acl/ZhaoWZNBCRYTH26}
& 7 & 1.0 & 1.0 & \cmark & \xmark \\

MMDR~\shortcite{DBLP:journals/corr/abs-2601-12346}
& -- & 2.8 & 1.0 & \cmark & \cmark \\

CRIT~\shortcite{DBLP:journals/corr/abs-2604-01634}
& -- & 6.5 & 2.7 & \xmark & \xmark \\

MC-Search~\shortcite{DBLP:journals/corr/abs-2603-00873}
& -- & 1.0 & 3.8 & \cmark & \cmark \\

VDR~\shortcite{DBLP:journals/corr/abs-2602-02185}
& -- & 1.0 & 1.0 & \cmark & \cmark \\
\midrule
\textbf{Sci-MMR (Ours)}
& \textbf{4} & \bfseries 9.0 & \bfseries 9.9 & \cmark & \cmark \\
\bottomrule
\end{tabular}%
}

\caption{Comparison with existing benchmarks. \textbf{Disc.} denotes the number of disciplines ($-$ indicates open-domain coverage). \textbf{Img.} and \textbf{Hop} denote per-task averages. \textbf{Agentic}: agent-based execution; \textbf{Process}: process-level evaluation.}
\label{tab:related-work}
\end{table}


\begin{figure*}[t]
      \centering
      \includegraphics[width=\textwidth]{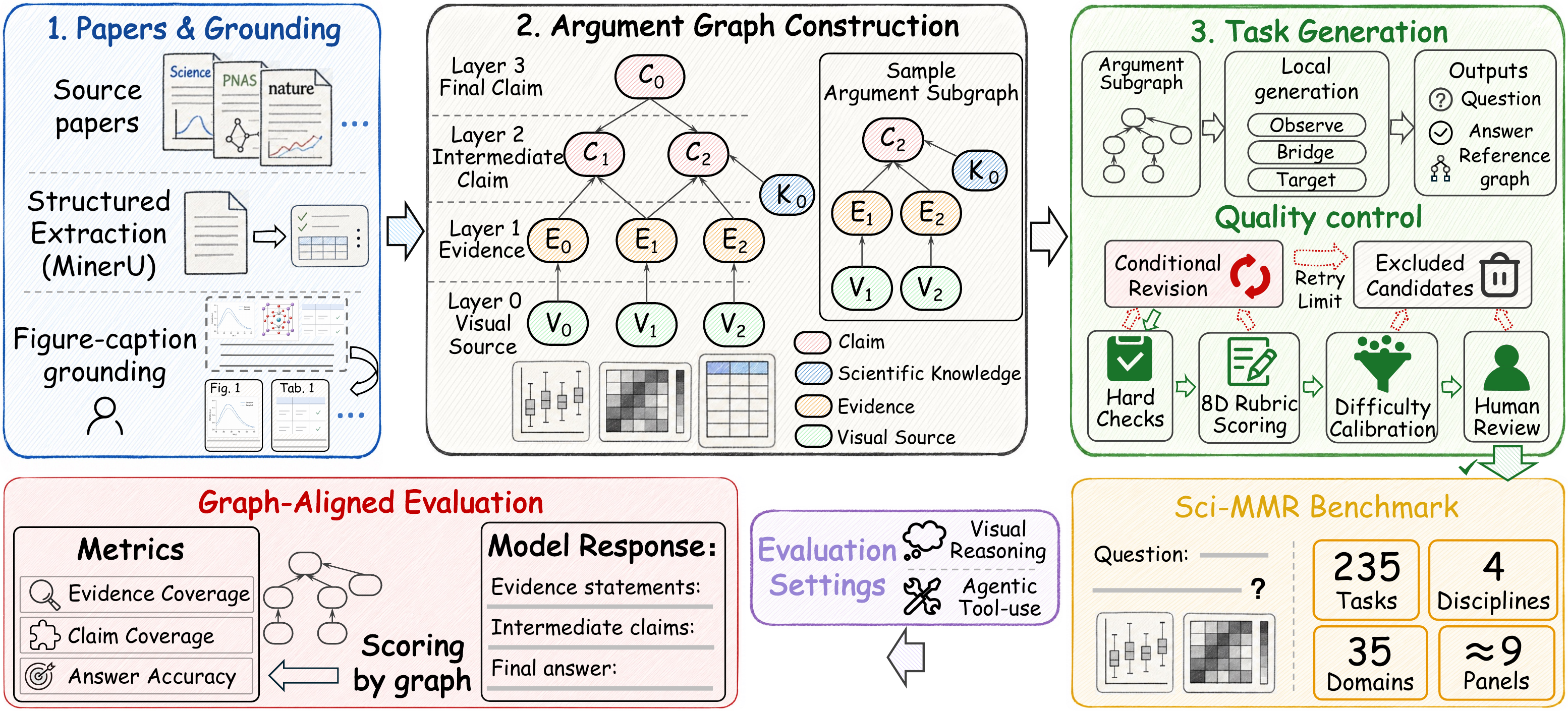}
      \caption{\textbf{Overview of \textsc{Sci-MMR}.} Scientific papers are grounded in visual evidence and represented as hierarchical Argument Graphs. Sampled argument subgraphs drive task generation and quality control, yielding 235 tasks spanning four disciplines and 35 domains, with an average of 9 figure panels per task. Model responses are evaluated using answer accuracy, evidence coverage, and claim coverage.}
      \label{fig:main}  
\end{figure*}

\section{\textsc{Sci-MMR} and Evaluation}

Scientific reasoning is inherently evidence-driven: conclusions are established by integrating multiple evidence items distributed across text, figures, and prior knowledge. Existing scientific multimodal benchmarks, however, primarily evaluate final answers without explicitly modeling the evidence dependencies underlying a scientific claim. Even benchmarks with multi-hop reasoning typically construct reasoning paths synthetically through graph sampling or retrieval expansion, rather than recovering the evidential argument presented in the original paper.

\textsc{Sci-MMR} addresses this limitation by reconstructing the evidence dependencies directly from peer-reviewed scientific publications. Each task is represented as an \emph{Argument Graph} that links scientific claims, evidence statements, citation-grounded knowledge, and supporting visual regions. This representation enables fine-grained evaluation of both evidence acquisition---whether a model retrieves the required evidence---and evidence integration---whether it correctly connects the acquired evidence to the target scientific claim.


\subsection{Argument Graph Formulation}

Scientific arguments are inherently hierarchical: visual evidence supports observations, which are recursively composed into increasingly abstract scientific claims~\citep{DBLP:conf/eacl/TeufelCM99,DBLP:conf/argmining/LauscherGP18,DBLP:conf/argmining/MoserM20}. We represent this structure as an \emph{Argument Graph}, a directed acyclic graph:
\[
\mathcal{G}=(V,E),
\]
where nodes \(V\) denote information at different abstraction levels and edges \(E\) encode support relationships.

\paragraph{Nodes.}
The node set is partitioned into four disjoint types:
\[
V = V_v \cup V_e \cup V_k \cup V_c.
\]

\begin{itemize}
    \item \(V_v\): \textbf{Visual source nodes} denote figure panels, tables, and other visual elements containing experimental information.
    
    \item \(V_e\): \textbf{Evidence nodes} represent evidence grounded in visual sources and link scientific statements to their supporting visual elements.
    
    \item \(V_k\): \textbf{Knowledge nodes} represent external scientific knowledge, methodological definitions, or prior findings grounded in the paper's cited references.
    
    \item \(V_c\): \textbf{Claim nodes} represent hierarchical scientific claims, from intermediate conclusions supported by evidence to the final scientific conclusion.
\end{itemize}

\paragraph{Edges.}
Each edge \((u,v)\in E\) denotes that node \(u\) directly supports node \(v\). Support relations are restricted to:
\[
E \subseteq \big(V_v \times V_e\big)
\;\cup\;
\big(V_e \times V_c\big)
\;\cup\;
\big(V_k \times V_c\big)
\;\cup\;
\big(V_c \times V_c\big),
\]
where \(V_v \times V_e\) grounds evidence in visual regions, \(V_e \times V_c\) links evidence to claims, \(V_k \times V_c\) captures dependencies on citation-grounded scientific knowledge, and \(V_c \times V_c\) composes lower-level claims into higher-level ones. Since \(\mathcal{G}\) is a DAG, every claim is supported by an acyclic chain of evidence and intermediate claims terminating at the task conclusion.

\paragraph{Root and task structure.}
Each graph has a unique root claim \(c^\ast\in V_c\), corresponding to the task's final scientific conclusion. Solving a task requires recovering the relevant evidence and knowledge nodes and integrating them through the argument graph to infer \(c^\ast\). We characterize each task by its \textbf{hop count}, \(\mathrm{Hop}=|E|\), \textbf{graph depth}, defined as the longest directed path terminating at \(c^\ast\), and \textbf{panel count}, \(|V_v|\), the number of distinct visual sources.

\subsection{Benchmark Construction and Composition}

We adopt a human-in-the-loop pipeline in which automated tools perform scalable extraction and graph construction, while domain experts intervene only where scientific judgment is required: evidence annotation and final verification.

\paragraph{Graph construction.}
We collect 800 papers published between 2020 and 2026 from leading venues including \textit{Nature}, \textit{Science}, \textit{Cell}, and \textit{PNAS}; the final benchmark is dominated by 2026 papers (Appendix~A.1). For each paper, MinerU~\citep{DBLP:journals/corr/abs-2409-18839} extracts figures, tables, and captions. An LLM constructs an initial Argument Graph linking visual sources, evidence, scientific knowledge, and claims, retrieving scientific knowledge from the paper's cited references when required. Domain experts then localize each evidence node to its supporting visual region, rewrite it as a conservative, directly observable statement, and verify the complete argument graph.

\paragraph{Task generation.}
From each verified Argument Graph, we sample a argument subgraph sufficient to derive a target claim. The sampled graph specifies the required observations, intermediate claims, and scientific knowledge, from which an LLM generates a natural multi-hop question without revealing the target conclusion. Reference answers are generated independently from the complete argument graph, presenting grounded observations, intermediate reasoning, and the final conclusion in dependency order with explicit figure and table attributions.
\paragraph{Automated quality control.} Candidate tasks first pass through two automated filters. Deterministic validation removes 54\% of tasks whose argument graph depth falls below a minimum threshold. The remaining tasks are evaluated using an eight-dimensional LLM-based quality rubric, filtering a further 17\%; failed tasks are revised and re-evaluated rather than discarded. The complete rubric is detailed in Supplementary Section B.

\paragraph{Human verification.} Automated filtering can only screen for structural and surface-level quality; it cannot verify whether a task's argument is scientifically sound. Domain experts therefore conduct a final manual review of every surviving task, confirming that each evidence statement is faithfully grounded in its visual region and that the reasoning chain genuinely supports the target claim. This is the last and only stage capable of catching tasks that pass automated checks yet fail scientifically. 

\begin{figure}[t]
    \centering
    \includegraphics[width=\columnwidth]{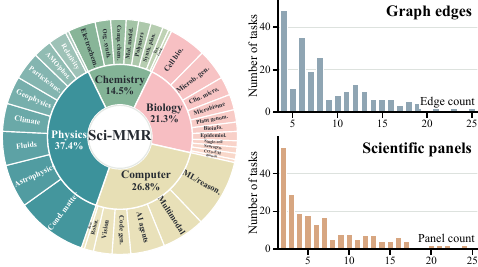}
    \caption{\textsc{Sci-MMR} composition and task complexity. Discipline--domain distribution and distributions of graph edges and figure panels per task.}
    \label{fig:discipline_domain_composition}
\end{figure}

This pipeline yields \textsc{Sci-MMR}, comprising 235 multi-hop reasoning tasks spanning 4 scientific disciplines and 35 domains, with an average of 9 figure panels per task. Figure~\ref{fig:discipline_domain_composition} summarizes the benchmark composition and task complexity, while Supplementary Secs.~A.2--A.3 provide additional dataset statistics.

\input{tables/main_results_by_difficulty}

\subsection{Evaluation Metrics}

Scientific reasoning proceeds from visual evidence to intermediate claims and ultimately to a scientific conclusion. Accordingly, \textsc{Sci-MMR} evaluates three complementary dimensions: (1) Answer Accuracy, (2) Evidence Coverage, and (3) Claim Coverage, measuring whether models recover the complete reasoning chain.

\paragraph{(1) Answer Accuracy.}
For a task $t \in \mathcal{T}$, let $r_t$ denote the model response and $y_t$ the reference conclusion. The task-level score is
\begin{equation}
A_t = \mathrm{match}_{\mathrm{ans}}(r_t, y_t),
\end{equation}
where $\mathrm{match}_{\mathrm{ans}}(r_t,y_t)=1$ if the final conclusion expressed in $r_t$ is semantically consistent with $y_t$, and $0$ otherwise. Overall Answer Accuracy is the dataset-level average:
\begin{equation}
\mathrm{Acc.} = \frac{1}{|\mathcal{T}|}\sum_{t\in\mathcal{T}} A_t.
\end{equation} 

\paragraph{(2) Evidence Coverage.}
For a task $t$ with reference evidence nodes $V_{e,t}$ and response $r_t$, the task-level score is
\begin{equation}
E_t = \frac{1}{|V_{e,t}|}\sum_{e_i \in V_{e,t}} m(e_i, r_t),
\end{equation}
where $m(e_i, r_t)=1$ if the response substantively expresses the scientific proposition represented by $e_i$, and $0$ otherwise. Overall Evidence Coverage is
\begin{equation}
\mathrm{E\mbox{-}Cov.} = \frac{1}{|\mathcal{T}|}\sum_{t\in\mathcal{T}} E_t.
\end{equation}
This metric evaluates evidence acquisition, i.e., whether models correctly transform visual evidence into grounded observations.

\paragraph{(3) Claim Coverage.}
Let $V_{c,t}^{\mathrm{int}} = V_{c,t}\setminus\{c_t^\ast\}$ denote the reference intermediate-claim nodes for task $t$, excluding the final root claim. The task-level score is
\begin{equation}
C_t = \frac{1}{|V_{c,t}^{\mathrm{int}}|}\sum_{c_i \in V_{c,t}^{\mathrm{int}}} m(c_i, r_t),
\end{equation}
defined only for tasks with $|V_{c,t}^{\mathrm{int}}|>0$. Overall Claim Coverage is the average over this subset:
\begin{equation}
\begin{gathered}
\mathrm{C\mbox{-}Cov.} = \frac{1}{|\mathcal{T}'|}\sum_{t\in\mathcal{T}'} C_t,\\
\mathcal{T}' = \{t \in \mathcal{T} : |V_{c,t}^{\mathrm{int}}| > 0\}.
\end{gathered}
\end{equation}

This metric evaluates evidence integration by measuring whether recovered evidence is composed into the intermediate claims required to support the final conclusion. 


\input{experiments}

\section{Limitations}

Our \textsc{Sci-MMR} provides a large-scale benchmark for evaluating evidence-grounded scientific reasoning and enables systematic analysis of evidence acquisition and evidence integration. Nevertheless, several limitations should be acknowledged.
First, an Argument Graph represents one plausible reconstruction of a paper's evidential structure rather than a unique ground truth. Evidence granularity, intermediate claims, and support relations may admit alternative yet valid interpretations. Although our LLM-assisted pipeline incorporates expert verification, some degree of construction subjectivity is unavoidable.
Second, Evidence Coverage and Claim Coverage evaluate the evidence and claims explicitly expressed in the final response rather than the model's internal reasoning process. In addition, although our evaluation demonstrates strong cross-judge agreement, LLM-based assessment may still be affected by semantic ambiguity, response style, and evaluator bias~\citep{DBLP:conf/nips/ZhengC00WZL0LXZ23,DBLP:conf/iclr/ZengYG0G024,DBLP:conf/emnlp/ChenWZHL25}.

\section{Conclusion}

We introduced \textsc{Sci-MMR}, a benchmark of 235 tasks for evaluating multi-step, evidence-grounded scientific reasoning. Its argument graphs connect visual sources, evidence, citation-grounded scientific knowledge, and claims, enabling evaluation beyond answer accuracy alone. Across eight frontier multimodal models, supplying gold evidence substantially improves performance, while equipping models with visual cropping tools yields only modest gains. Errors persist even when evidence is explicitly provided, revealing challenges in both evidence acquisition and evidence integration. These findings show that answer accuracy alone can overestimate scientific reasoning reliability, motivating evaluation that traces the complete, structured evidence behind a conclusion.

Looking ahead, a natural next step is to extend \textsc{Sci-MMR} from within-paper argument graphs to cross-paper argument modeling, enabling evaluation of whether systems can connect complementary or conflicting evidence and claims across scientific articles.

\bibliography{fdnlp_paper}

\appendix
\input{fdnlp_appendix_body.tex}


\end{document}

%% file: tables/main_results_by_difficulty.tex
\begin{table*}[t]
\centering
{\small
\setlength{\tabcolsep}{0.7mm}
\begin{tabular}{@{}l@{\hspace{3.0mm}}rrr@{\hspace{6.0mm}}rrr@{\hspace{6.0mm}}rrr@{\hspace{6.0mm}}rrr@{}}
\toprule
\textbf{Model} & \multicolumn{3}{c}{\textbf{Easy} ($n=62$)} & \multicolumn{3}{c}{\textbf{Medium} ($n=92$)} & \multicolumn{3}{c}{\textbf{Hard} ($n=81$)} & \multicolumn{3}{c}{\textbf{Overall} ($n=235$)} \\
\cmidrule(lr){2-4}\cmidrule(lr){5-7}\cmidrule(lr){8-10}\cmidrule(l){11-13}
& Acc. & E-Cov. & C-Cov. & Acc. & E-Cov. & C-Cov. & Acc. & E-Cov. & C-Cov. & Acc. & E-Cov. & C-Cov. \\
\midrule
\multicolumn{13}{c}{\emph{Direct Visual Reasoning}} \\
\cmidrule(lr){1-13}
GPT-5.5 & \textbf{96.8} & 66.0 & 56.3 & \textbf{81.5} & \underline{60.7} & 46.4 & \textbf{29.6} & \underline{50.3} & 36.5 & \textbf{67.7} & 58.5 & 46.1 \\
Claude Opus 4.8 & \underline{95.2} & \underline{75.7} & \textbf{64.6} & \underline{66.3} & 54.8 & \textbf{51.4} & \underline{11.1} & \textbf{51.8} & \underline{39.7} & \underline{54.9} & \underline{59.3} & \underline{51.5} \\
Kimi K2.7 & 93.5 & 59.1 & 39.7 & 56.5 & 51.1 & 42.1 & \underline{11.1} & 44.6 & 27.0 & 50.6 & 50.9 & 36.7 \\
Gemini 3.1 Pro & 90.3 & 58.0 & 43.1 & 58.7 & 42.1 & 40.6 & 9.9 & 35.0 & 34.6 & 50.2 & 43.8 & 39.4 \\
MiniMax-M3 & 93.5 & \textbf{77.8} & \underline{57.9} & 48.9 & \textbf{60.8} & \underline{50.6} & 4.9 & 49.9 & \textbf{47.5} & 45.5 & \textbf{61.5} & \textbf{51.7} \\
GLM-5V Turbo & 83.9 & 58.9 & 40.6 & 41.3 & 45.6 & 32.3 & 8.6 & 41.3 & 29.9 & 41.3 & 47.6 & 33.9 \\
Qwen3.7 Plus & 71.0 & 57.9 & 43.9 & 28.3 & 40.3 & 31.3 & 4.9 & 40.7 & 30.4 & 31.5 & 45.1 & 34.6 \\
Intern-S2 & 71.0 & 57.5 & 38.8 & 16.3 & 38.1 & 33.6 & 4.9 & 33.9 & 21.1 & 26.8 & 41.8 & 31.2 \\
\cmidrule(lr){1-13}
\emph{Average} & 86.9 & 63.9 & 48.1 & 49.7 & 49.2 & 41.0 & 10.6 & 43.4 & 33.3 & 46.1 & 51.1 & 40.6 \\
\midrule
\multicolumn{13}{c}{\emph{Agentic Tool-Use}} \\
\cmidrule(lr){1-13}
GPT-5.5 & \textbf{100.0} & 73.8 & 63.4 & \textbf{84.8} & 62.6 & 51.9 & \textbf{29.6} & \underline{57.2} & \textbf{46.3} & \textbf{69.8} & 63.7 & 53.4 \\
Claude Opus 4.8 & \underline{93.5} & \underline{79.5} & \textbf{68.3} & \underline{72.8} & \underline{67.4} & \underline{52.0} & 19.8 & \textbf{59.7} & 43.4 & \underline{60.0} & \textbf{67.9} & \underline{54.0} \\
Kimi K2.7 & \underline{93.5} & 67.9 & 48.8 & 63.0 & 54.0 & 42.8 & 9.9 & 48.0 & 39.7 & 52.8 & 55.6 & 43.5 \\
Gemini 3.1 Pro & 88.7 & 68.2 & 49.4 & 54.3 & 55.0 & \textbf{56.7} & \underline{22.2} & 48.3 & 43.4 & 52.3 & 56.2 & 50.5 \\
MiniMax-M3 & 83.9 & \textbf{80.3} & \underline{67.1} & 58.7 & \textbf{70.3} & 51.0 & 16.0 & 54.6 & \textbf{46.3} & 50.6 & \underline{67.5} & \textbf{54.1} \\
GLM-5V Turbo & 82.3 & 68.9 & 47.8 & 46.7 & 56.3 & 42.3 & 12.3 & 49.9 & \underline{43.6} & 44.3 & 57.4 & 44.3 \\
Qwen3.7 Plus & 71.0 & 65.4 & 49.0 & 44.6 & 53.4 & 36.0 & 13.6 & 48.2 & 39.0 & 40.9 & 54.8 & 40.6 \\
Intern-S2  & 71.0 & 44.2 & 26.9 & 28.3 & 28.3 & 17.1 & 12.3 & 28.7 & 24.5 & 34.0 & 32.6 & 22.1 \\
\cmidrule(lr){1-13}
\emph{Average} & 85.5 & 68.5 & 52.6 & 56.7 & 55.9 & 43.7 & 17.0 & 49.3 & 40.8 & 50.6 & 57.0 & 45.3 \\
\bottomrule
\end{tabular}
}
\par\vspace{1.2mm}
\includegraphics[width=0.92\textwidth]{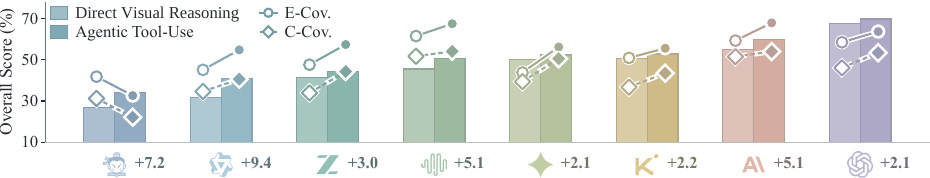}
\caption{Direct Visual Reasoning and Agentic Tool-Use performance by task difficulty and overall. Scores are percentages; overall results are unweighted averages. Bold and underlined values indicate the best and second-best performance, respectively. The lower panel compares overall Answer Accuracy (Acc.), Evidence Coverage (E-Cov.), and Claim Coverage (C-Cov.); hollow and filled markers denote Direct Visual Reasoning and Agentic Tool-Use, respectively.}
\label{tab:main_results_by_difficulty}
\end{table*}

%% file: experiments.tex
\section{Experiments}
\label{sec:experiments}

We structure our experiments as a progressively constrained diagnostic evaluation that isolates failures in evidence acquisition, evidence integration, and final-answer reasoning.

\paragraph{Evaluation Models.}
We evaluate eight frontier multimodal models on \textsc{Sci-MMR}: five proprietary models---GPT-5.5~\citep{openai2026gpt55}, Claude Opus~4.8~\citep{anthropic2026opus48}, Gemini~3.1 Pro~\citep{google2026gemini31pro}, GLM-5V-Turbo~\citep{glmVTeam2026glm5vturbo}, and Qwen3.7-Plus~\citep{qwen2026qwen37plus}---and three open-weight models: Kimi K2.7 Code~\citep{moonshot2026kimik27code}, MiniMax M3~\citep{minimax2026minimaxm3}, and Intern-S2-Preview-FP8~\citep{internlm2026interns2preview}.

\paragraph{Evaluation Settings.}
We evaluate all models under four progressively constrained settings that isolate different stages of evidence-grounded scientific reasoning.

\begin{itemize}
    \item \textbf{Caption-only.} Models receive only figure captions, measuring the extent to which tasks can be solved from textual context alone.
    
    \item \textbf{Direct Visual Reasoning.} Models receive the original question and scientific figures, requiring them to acquire visual evidence and derive the final conclusion autonomously.
    
    \item \textbf{Evidence-Hint Reasoning.} Models additionally receive expert-annotated evidence statements, isolating evidence integration by removing the need for visual evidence acquisition.
    
    \item \textbf{Agentic Tool-Use.} Models are equipped with visual tools for cropping and magnifying regions of interest while preserving the original context, evaluating whether interactive evidence acquisition improves scientific reasoning.
\end{itemize}

\paragraph{Metrics.}
We evaluate free-form responses using three complementary metrics. \textit{Answer Accuracy} (Acc.) measures whether the model reaches the correct scientific conclusion. \textit{Evidence Coverage} (E-Cov.) measures whether the response recovers the required visual evidence. \textit{Claim Coverage} (C-Cov.) measures whether the response recovers the intermediate claims connecting evidence to the final conclusion. Together, these metrics distinguish answer correctness from evidence acquisition and evidence integration. All responses are evaluated by Gemini 3.5 Flash~\citep{google2026gemini35flash} using a unified rubric for answer correctness and graph coverage. Human agreement and cross-judge validation are reported in Supplementary Sec.~C.3.



\subsection{Main Results}
\label{sec:main_results}

Table~\ref{tab:main_results_by_difficulty} reports Answer Accuracy, Evidence Coverage, and Claim Coverage under Direct Visual Reasoning and Agentic Tool-Use across three task difficulty tiers.

\noindent\textbf{Direct Visual Reasoning.}\;
GPT-5.5 achieves the highest overall accuracy at 67.7\%, followed by Claude Opus~4.8 at 54.9\% and Kimi K2.7 at 50.6\%, while Intern-S2 ranks last at 26.8\%. Performance deteriorates sharply with task difficulty for every model, but the extent of degradation varies substantially. GPT-5.5 drops from 96.8\% on easy tasks to 29.6\% on hard tasks, yet remains the strongest model on the hardest subset. In contrast, MiniMax-M3 declines from 93.5\% to 4.9\%, exhibiting the largest degradation of any model. These results suggest that strong performance on easier tasks does not reliably translate to complex multi-hop scientific reasoning, where deeper evidence integration becomes the dominant bottleneck.

\noindent\textbf{Agentic Tool-Use.}\;
Equipping models with visual tools consistently improves performance, although the gains remain modest. Claude Opus~4.8 achieves the largest improvement at 5.1 points, whereas Gemini 3.1 Pro and Kimi K2.7 improve by only 2.1 points each. GPT-5.5 remains the strongest model overall at 69.8\% and across all three difficulty tiers, indicating that interactive evidence acquisition alone does not close the gap on complex scientific reasoning.


\noindent\textbf{Evidence and Claim Coverage.}\;
Table~\ref{tab:main_results_by_difficulty} also reports Evidence Coverage and Claim Coverage alongside Answer Accuracy. Across both evaluation settings, the three metrics exhibit different rankings, indicating that answer correctness, evidence acquisition, and evidence integration capture distinct aspects of scientific reasoning. A representative example is MiniMax-M3, which achieves the highest overall Evidence Coverage of any model under Direct Visual Reasoning at 61.5\% and the highest Claim Coverage at 51.7\%, yet ranks only fifth in Answer Accuracy at 45.5\%, trailing GPT-5.5, Claude Opus~4.8, Kimi K2.7, and Gemini 3.1 Pro. Similar dissociations are observed within individual models, as detailed in Supplementary Section D.3, demonstrating that higher evidence or claim coverage does not necessarily translate into correct scientific conclusions.

\begin{figure}[!htbp]
    \centering
    \includegraphics[width=\columnwidth]{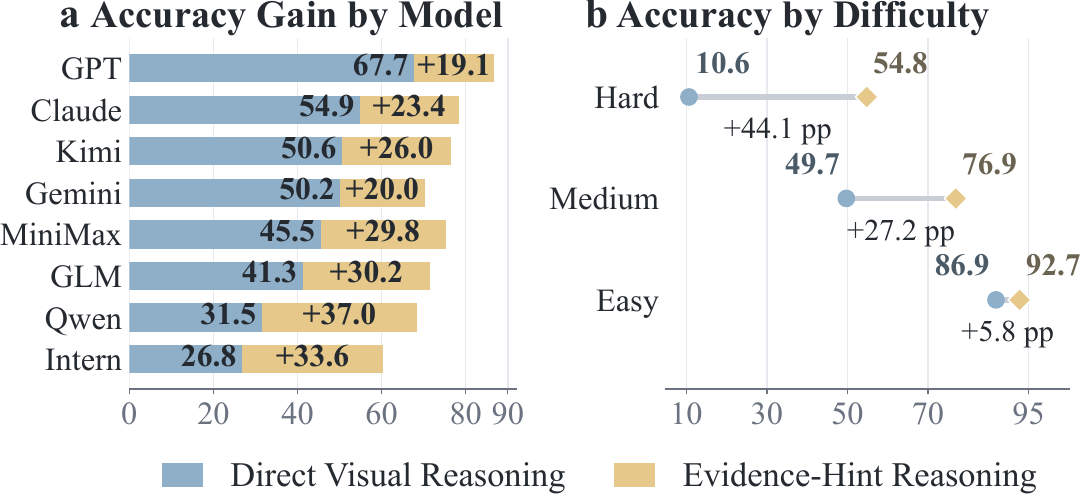}
    \caption{Accuracy gains from Direct Visual Reasoning to Evidence-Hint Reasoning by model (a) and difficulty (b).}   
    \label{fig:evidence_hint_intervention}
\end{figure}  

Simple-task accuracy, tool-assisted gains, and evidence coverage each expose a limitation in current evaluation, but none pinpoint its cause. This leaves a deeper question open: do frontier multimodal agents genuinely perform multi-step, evidence-grounded scientific reasoning, or merely reach correct answers through other means? To this end, we propose three research questions targeting successive stages of the reasoning pipeline. \textbf{RQ1} asks whether evidence availability alone is sufficient for reliable reasoning. \textbf{RQ2} asks whether autonomous evidence acquisition is the primary bottleneck. \textbf{RQ3} asks specifically where the remaining errors originate. Together, \textbf{RQ1}--\textbf{RQ3} trace the full pipeline from evidence acquisition to grounded reasoning, moving from establishing that a gap exists to locating its source and characterizing its precise manifestation.

\paragraph{RQ1:\ Is Evidence Availability Sufficient for Reliable Scientific Reasoning?}
Figure~\ref{fig:evidence_hint_intervention} shows that supplying gold evidence statements under Evidence-Hint Reasoning raises accuracy for every model, but the gains fall well short of closing the gap to reliable performance. Averaged across models, accuracy rises by 27.4 points, from 46.1\% to 73.5\%, yet mean accuracy on hard tasks reaches only 54.8\%, still far from ceiling. Models that scored lowest under Direct Visual Reasoning tend to gain the most: Intern-S2, the weakest model at 26.8\%, gains 33.6 points, while GPT-5.5, the strongest at 67.7\%, gains only 19.1 points, consistent with acquisition rather than reasoning capacity being their binding constraint. The pattern also holds across difficulty tiers: hard tasks gain the most at 44.1 points, compared to 27.2 points on medium tasks and only 5.8 points on easy tasks, indicating that evidence availability disproportionately helps on harder tasks without eliminating their difficulty. Evidence availability is therefore necessary but not sufficient: even when the complete evidence set is handed to a model, integrating it into a correct scientific conclusion remains a substantial, unresolved challenge. 



\begin{figure}[htbp]
    \centering
    \includegraphics[width=\columnwidth]{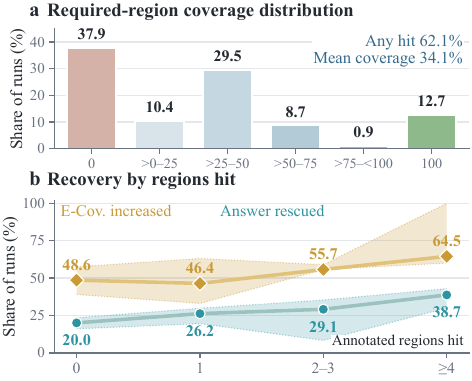}
    \caption{Crop localization and recovery at IoU $=0.5$. (a) Required-region coverage among crop-called runs. (b) Shares of initially incorrect, low-E-Cov. runs with increased E-Cov. or a rescued answer, grouped by annotated regions hit; bands show cross-model IQRs.}
    \label{fig:crop_localization_rescue}
\end{figure}

\paragraph{RQ2:\ Is Autonomous Evidence Acquisition the Primary Bottleneck?}

Figure~\ref{fig:crop_localization_rescue} examines whether models can autonomously locate and recover the evidence required for a task. Among crop-called runs, models hit at least one annotated region in 62.1\% of cases, but the distribution of coverage is heavily skewed: 37.9\% of runs hit no required region at all, while only 12.7\% achieve complete coverage, yielding a mean coverage of just 34.1\%. This confirms that acquisition failure is widespread and severe. However, acquisition alone does not fully explain the accuracy gap. In panel (b), 46.4\%--64.5\% of runs improve in E-Cov., with the strongest recovery in the upper-hit bins, while answer rescue rises from 20.0\% to 38.7\%. Greater region access is therefore associated with better recovery. However, localization alone remains insufficient: even with at least four hits, only 38.7\% of answers are rescued. This gap indicates that locating relevant evidence does not guarantee its correct extraction and integration, motivating the error analysis in RQ3. We additionally assess sensitivity to the IoU criterion using thresholds of 0.3 and 0.7 around the default of 0.5 (Supplementary Sec.~F.4).

\paragraph{RQ3:\ When Reasoning Fails, What Specifically Goes Wrong?}

\begin{figure}[!htbp]
    \centering
    \includegraphics[width=\columnwidth]{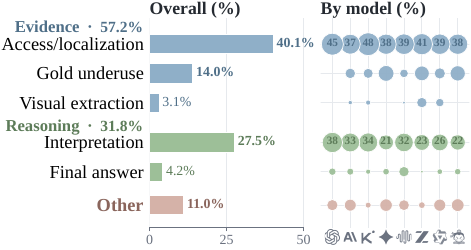}
    \caption{Error patterns among incorrect Direct Visual Reasoning outputs. The left panel shows the overall distribution, and the right panel shows the distribution for each model. Bubble size indicates the percentage of a model's errors.}
    \label{fig:cross_setting_error_patterns}
\end{figure}

RQ2 shows that autonomous evidence acquisition is a dominant but incomplete explanation for the accuracy--coverage gap, leaving open what accounts for the remaining errors. Figure~\ref{fig:cross_setting_error_patterns} addresses this by decomposing all incorrect Direct Visual Reasoning outputs into six failure subtypes using a cross-setting routing procedure. Evidence-related failures account for 57.2\% of errors overall, dominated by access and localization failures at 40.1\%, with gold underuse at 14.0\% and visual extraction errors at 3.1\% contributing smaller shares. Reasoning-related failures account for a further 31.8\%, split between interpretation failures at 27.5\% and final-answer mismatches at 4.2\%; the remaining 11.0\% fall outside these categories. This breakdown is consistent across nearly all eight evaluated models: access and localization failures remain the largest single subtype for every model, ranging from 37\% for Claude Opus~4.8 to 48\% for Kimi K2.7, while interpretation failures form the second-largest category for most models. Taken together with RQ2, this fine-grained attribution confirms that current multimodal models fail along two largely distinct axes -- acquiring complete evidence from raw visuals, and correctly interpreting evidence once retrieved -- rather than along a single, dominant failure mode. 

%% file: fdnlp_appendix_body.tex
\captionsetup{skip=2pt}
\setlength{\textfloatsep}{5pt plus 1pt minus 1pt}
\setlength{\dbltextfloatsep}{5pt plus 1pt minus 1pt}
\setlength{\floatsep}{4pt plus 1pt minus 1pt}
\setlength{\dblfloatsep}{4pt plus 1pt minus 1pt}
\setlength{\intextsep}{5pt plus 1pt minus 1pt}
\setcounter{topnumber}{4}
\setcounter{dbltopnumber}{3}
\setcounter{totalnumber}{6}
\renewcommand{\topfraction}{0.92}
\renewcommand{\dbltopfraction}{0.92}
\renewcommand{\textfraction}{0.08}
\renewcommand{\floatpagefraction}{0.90}
\renewcommand{\dblfloatpagefraction}{0.90}

\newtcblisting{PromptCard}[2][]{%
    enhanced,
    breakable,
    listing only,
    colback=white,
    colframe=black,
    boxrule=0.7pt,
    arc=1.2mm,
    outer arc=1.2mm,
    left=0.55mm,
    right=0.55mm,
    top=0.7mm,
    bottom=0.55mm,
    title={#2},
    coltitle=white,
    fonttitle=\sffamily\bfseries\scriptsize,
    attach boxed title to top left={%
        xshift=2.5mm,
        yshift*=-\tcboxedtitleheight/2
    },
    boxed title style={%
        colback=black,
        colframe=black,
        boxrule=0pt,
        arc=0.8mm,
        left=0.75mm,
        right=0.75mm,
        top=0.3mm,
        bottom=0.3mm
    },
    listing options={%
        numbers=none,
        xleftmargin=0pt,
        basicstyle=\ttfamily\fontsize{6.0}{7.1}\selectfont,
        breaklines=true,
        breakatwhitespace=false,
        breakautoindent=false,
        breakindent=0pt,
        literate={_}{{\_\allowbreak}}1,
        columns=fullflexible,
        keepspaces=true,
        showstringspaces=false
    },
    before skip=1.2pt plus 0.3pt,
    after skip=1.2pt plus 0.3pt,
    #1
}
\makeatletter
\newcommand{\InlineTableBegin}[1][]{%
    \par\smallskip
    \noindent\begin{minipage}{\columnwidth}
    \centering
    \captionsetup{type=table}%
}
\newcommand{\InlineTableEnd}{%
    \end{minipage}
    \par\smallskip
}
\newcommand{\SingleColumnTableInput}[1]{%
    \begingroup
    \footnotesize
    \setlength{\textwidth}{\columnwidth}%
    \setlength{\linewidth}{\columnwidth}%
    \setlength{\tabcolsep}{2.5pt}%
    \renewcommand{\arraystretch}{1.04}%
    \let\table\InlineTableBegin
    \let\endtable\InlineTableEnd
    \expandafter\let\csname table*\endcsname\InlineTableBegin
    \expandafter\let\csname endtable*\endcsname\InlineTableEnd
    \input{#1}%
    \endgroup
}
\newcommand{\WideTableInput}[1]{%
    \begingroup
    \scriptsize
    \setlength{\tabcolsep}{2pt}%
    \renewcommand{\arraystretch}{1.02}%
    \input{#1}%
    \endgroup
}
\makeatother
\setcounter{secnumdepth}{2}
\section*{Appendix}

\section{Benchmark Composition and Audit}
\label{sec:supp_benchmark_composition}

The \textsc{Sci-MMR} benchmark contains 235 questions spanning biology, chemistry, computer science, and physics. We first report the source-pool and benchmark composition, then characterize the structural and visual demands of individual questions, provide an audited graph example, and document expert verification.

\subsection{Temporal, Disciplinary, and Domain Coverage}
\label{sec:supp_domain_coverage}

The construction pool contains 800 deduplicated papers: 50 from the \textit{Nature} family, 40 from the \textit{Science} family, 30 from the \textit{Cell} family, 60 from \textit{PNAS}, and 620 from other sources. MinerU parsing succeeds for 700 papers, argument graphs are available for 475, and 300 papers contribute candidate tasks. The task-level validation flow follows the two automated rates reported in the main text: deterministic validation removes 54\% of the candidate-task inventory, and the subsequent rubric stage filters a further 17\%. Failed candidates are revised and re-evaluated. The final expert-review cohort contains 260 task IDs; 25 are rejected and 235 are retained in the benchmark. The final 235 tasks come from 170 papers; 119 papers contribute one task, 39 contribute two, 10 contribute three, and two contribute four.

For temporal coverage, we record each final task's source paper by its latest release or version year, which may differ from its original publication year. The 235 benchmark questions are not uniformly distributed across 2020--2026: the respective counts are 2, 2, 3, 16, 46, 38, and 128. Thus, 128 questions (54.5\%) use papers whose latest release or version is dated 2026.

Table~\ref{tab:supp_task_flow} summarizes the task-level construction flow; paper-level counts are upstream inventory counts.

\begin{table*}[!t]
    \centering
    \scriptsize
    \setlength{\tabcolsep}{4pt}
    \begin{tabularx}{\textwidth}{@{}p{0.28\textwidth}r X@{}}
        \toprule
        \textbf{Stage or event} & \textbf{Reported outcome} & \textbf{Identifier-preserving disposition and reason} \\
        \midrule
        Deduplicated paper pool & 800 papers & Upstream source inventory; no task IDs yet. \\
        MinerU parsing available & 700 papers & Papers retained for downstream graph construction. \\
        Argument graphs available & 475 papers & Papers with a usable graph. \\
        Candidate-task generation & 300 papers & Candidate tasks receive an immutable identifier before validation. \\
        Deterministic validation & 54\% removed & Depth $<3$ or invalid graph; surviving IDs retain their identity. \\
        Rubric and hard checks & 17\% further filtered & Failed candidates are revised and re-evaluated. The 66 below-threshold flags, 140 rewrite triggers, and 130 accepted rewrites are overlapping event counts. \\
        Expert review & 260 tasks & Two independent experts check each surviving ID; revisions retain the task ID through adjudication. \\
        Released benchmark & 235 tasks & 25 expert-review rejections are terminal dispositions; the retained IDs form the released benchmark. \\
        \bottomrule
    \end{tabularx}
    \caption{Task construction and quality-control flow. Stage rates use the single candidate-task denominator reported in the main text. Immutable IDs persist through rewrites and adjudication; the 54\% and 17\% entries are rates, while rewrite flags are overlapping events.}
    \label{tab:supp_task_flow}
\end{table*}

Table~\ref{tab:supp_domain_counts} reports the complete discipline--primary-domain partition using the intrinsic sample taxonomy. Physics contributes 88 questions (37.4\%), followed by computer science with 63 (26.8\%), biology with 50 (21.3\%), and chemistry with 34 (14.5\%). Within biology, the largest domains are microbial cell biology and physiology (12), microbial genomics and metagenomics (10), and clinical microbiology and infectious disease (6). Chemistry is led by electrochemistry and electrocatalysis (9), followed by organic synthesis and mechanism and quantum and computational chemistry (5 each). Computer science is led by machine learning and reasoning (22), language models and multimodal systems (14), and AI agents and tool-use systems (11). Physics is led by condensed matter and materials physics (23), astrophysics and cosmology (14), and fluids and plasma physics (11). The remaining domains contain between one and nine questions each. The resulting mix emphasizes physics and computer science while maintaining broad topical coverage: the largest primary domain contains 23 questions, or 9.8\% of the benchmark.

\begin{table}[!t]
    \centering
    \scriptsize
    \setlength{\tabcolsep}{2.5pt}
    \renewcommand{\arraystretch}{1.04}
    \begin{tabular}{@{}p{0.28\columnwidth}p{0.66\columnwidth}@{}}
        \toprule
        \textbf{Discipline} & \textbf{Primary domain} \\
        \midrule
        Biology (50) & microbial cell biology \& physiology (12) \\
        & microbial genomics \& metagenomics (10) \\
        & clinical microbiology \& infectious disease (6) \\
        & microbial ecology \& microbiome (5) \\
        & plant genomics \& crop biology (4) \\
        & computational biology \& bioinformatics (3) \\
        & public health microbiology \& epidemiology (3) \\
        & immunology \& single-cell immune profiling (2) \\
        & neurogenomics \& functional genomics (2) \\
        & structural biology \& cryo-electron microscopy (2) \\
        & biotechnology \& applied microbiology (1) \\
        \addlinespace[1pt]
        Chemistry (34) & electrochemistry \& electrocatalysis (9) \\
        & organic synthesis \& mechanism (5) \\
        & quantum \& computational chemistry (5) \\
        & molecular modeling \& cheminformatics (4) \\
        & polymer \& macromolecular chemistry (4) \\
        & reaction informatics \& synthesis planning (4) \\
        & toxicology \& environmental chemistry (2) \\
        & catalysis \& energy conversion (1) \\
        \addlinespace[1pt]
        Computer science (63) & machine learning \& reasoning (22) \\
        & language models \& multimodal systems (14) \\
        & AI agents \& tool-use systems (11) \\
        & code generation \& software engineering (6) \\
        & computer vision \& visual computing (6) \\
        & embodied AI \& robotics (3) \\
        & retrieval \& search systems (1) \\
        \addlinespace[1pt]
        Physics (88) & condensed matter \& materials physics (23) \\
        & astrophysics \& cosmology (14) \\
        & fluids \& plasma physics (11) \\
        & climate \& atmospheric physics (9) \\
        & geophysics \& seismology (9) \\
        & particle \& nuclear physics (9) \\
        & AMO, photonics \& spectroscopy (6) \\
        & gravitation \& relativity (6) \\
        & quantum information \& simulation (1) \\
        \bottomrule
    \end{tabular}
    \caption{Question counts for all 35 primary domains. Counts in parentheses sum to the discipline totals shown in the column headers.}
    \label{tab:supp_domain_counts}
\end{table}

\subsection{Structural and Visual Complexity}
\label{sec:supp_structural_visual_complexity}

\begin{figure*}[!t]
    \centering
    \includegraphics[width=\textwidth]{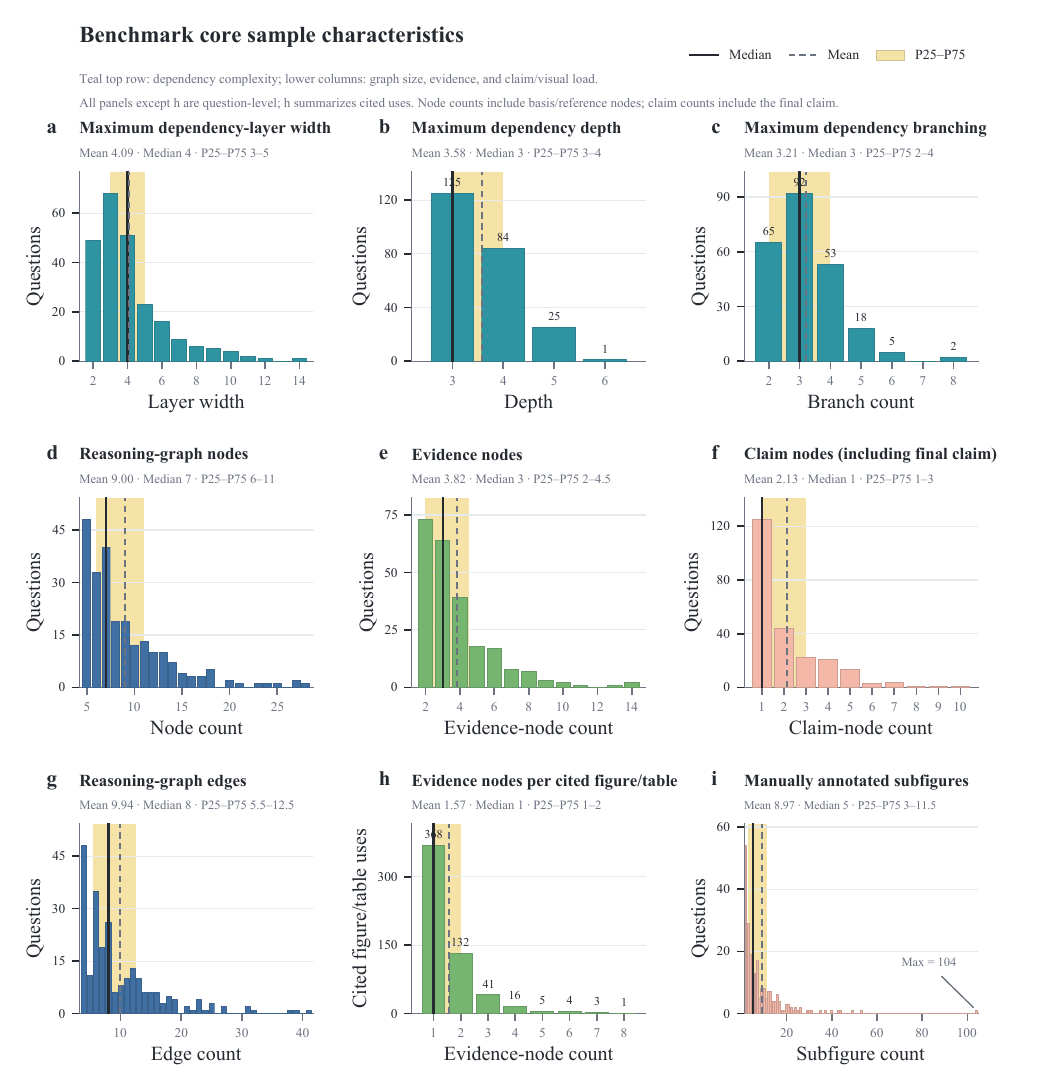}
    \caption{Core structural and visual characteristics of the benchmark. All panels except (h) summarize 235 questions; panel (h) summarizes 570 cited figure/table uses. The teal top row, panels (a)--(c), shows dependency width, depth, and branching. In the lower two rows, the blue column, panels (d) and (g), shows reasoning-graph nodes and edges; the green column, panels (e) and (h), shows question-level evidence nodes and evidence nodes per cited use; and the pink column, panels (f) and (i), shows claim nodes and manually annotated subfigures. Solid and dashed lines mark the median and mean, respectively, and cream spans show the interquartile range.}
    \label{fig:supp_benchmark_core_metrics}
\end{figure*}

Figure~\ref{fig:supp_benchmark_core_metrics} places related quantities next to one another: dependency complexity occupies the top row, while the lower rows form columns for graph size, evidence composition, and claim/visual load. Layer width is the maximum number of nodes at any dependency level, depth is the length of the longest dependency chain, and branching is the maximum number of direct dependencies from one node. These distributions are compact but nontrivial, with median layer width four, depth three, and branching three. The deterministic structural check requires a minimum dependency depth of 3. In the final 235-task benchmark, the depth counts are 125, 84, 25, and 1 at depths 3, 4, 5, and 6, respectively. The graph-size column gives medians of seven nodes and eight edges. The evidence column shows a median of three evidence nodes per question, while 500 of the 570 cited figure/table uses (87.7\%) support no more than two evidence nodes. The claim/visual column shows a median of one claim node and five manually annotated subfigures per question; the subfigure count is more heterogeneous, with a mean of 8.97 and a maximum of 104. Thus, most local visual references have focused evidential roles even though a complete question may combine several references within a larger reasoning graph.

\subsection{Audited Argument-Graph Example}
\label{sec:supp_argument_graph_example}

\begin{figure*}[!t]
    \centering
    \includegraphics[width=\textwidth,height=0.76\textheight,keepaspectratio]{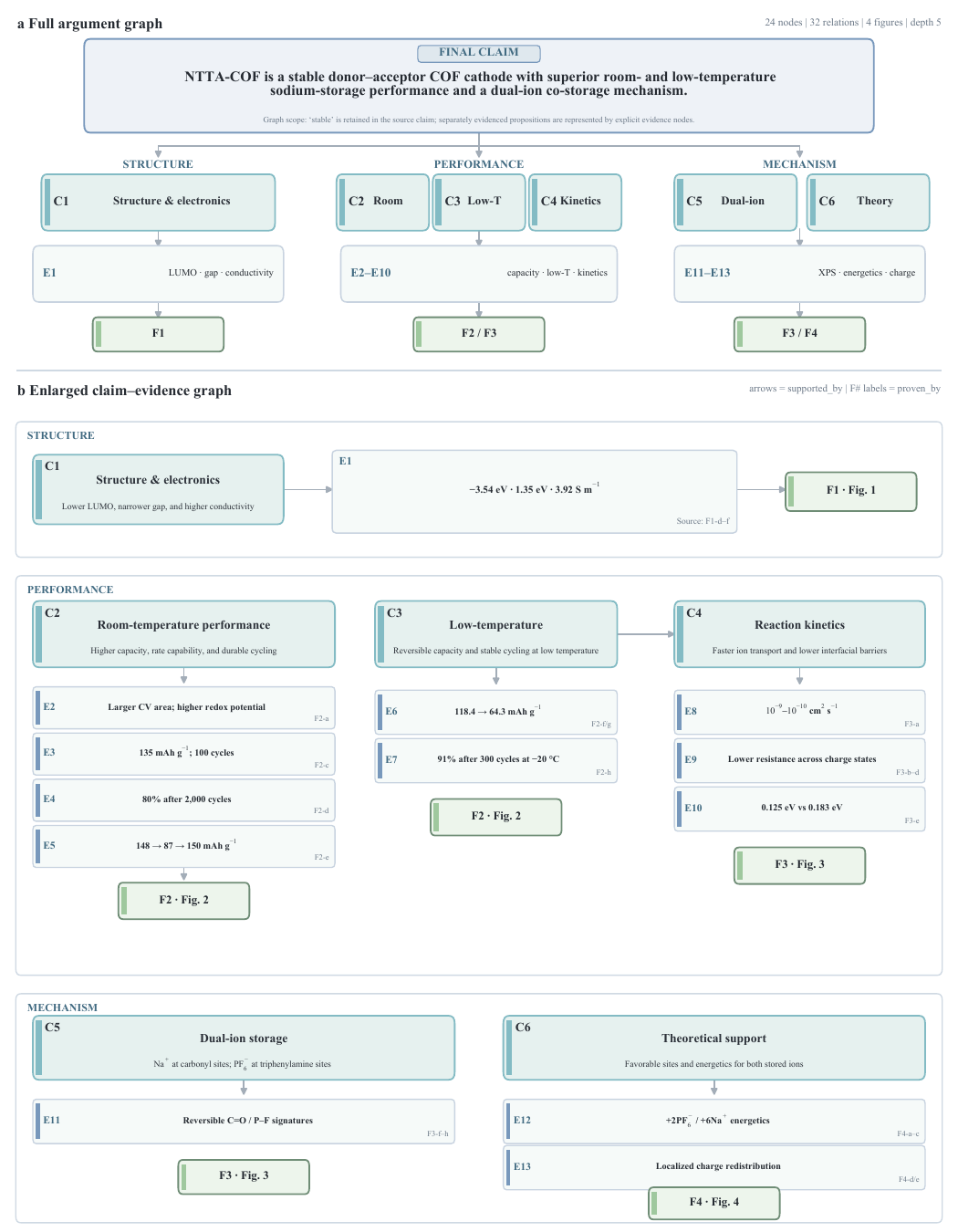}
    \caption{Audited full argument graph for one benchmark question. Panel (a) summarizes the final claim and its structural, performance, and mechanism branches. Panel (b) expands all intermediate claims and evidence nodes. Arrows link each conclusion to the evidence or lower-level claim that supports it, and each F\# label identifies the source panel for an evidence node. The graph contains seven claim nodes, 13 evidence nodes, four source-figure nodes, and 32 relations. The source claim's use of ``stable'' is retained in the claim text, while the separately evidenced propositions are represented by explicit evidence nodes.}
    \label{fig:supp_full_argument_graph}
\end{figure*}

\begin{figure*}[!t]
    \centering
    \includegraphics[width=\textwidth,height=0.76\textheight,keepaspectratio]{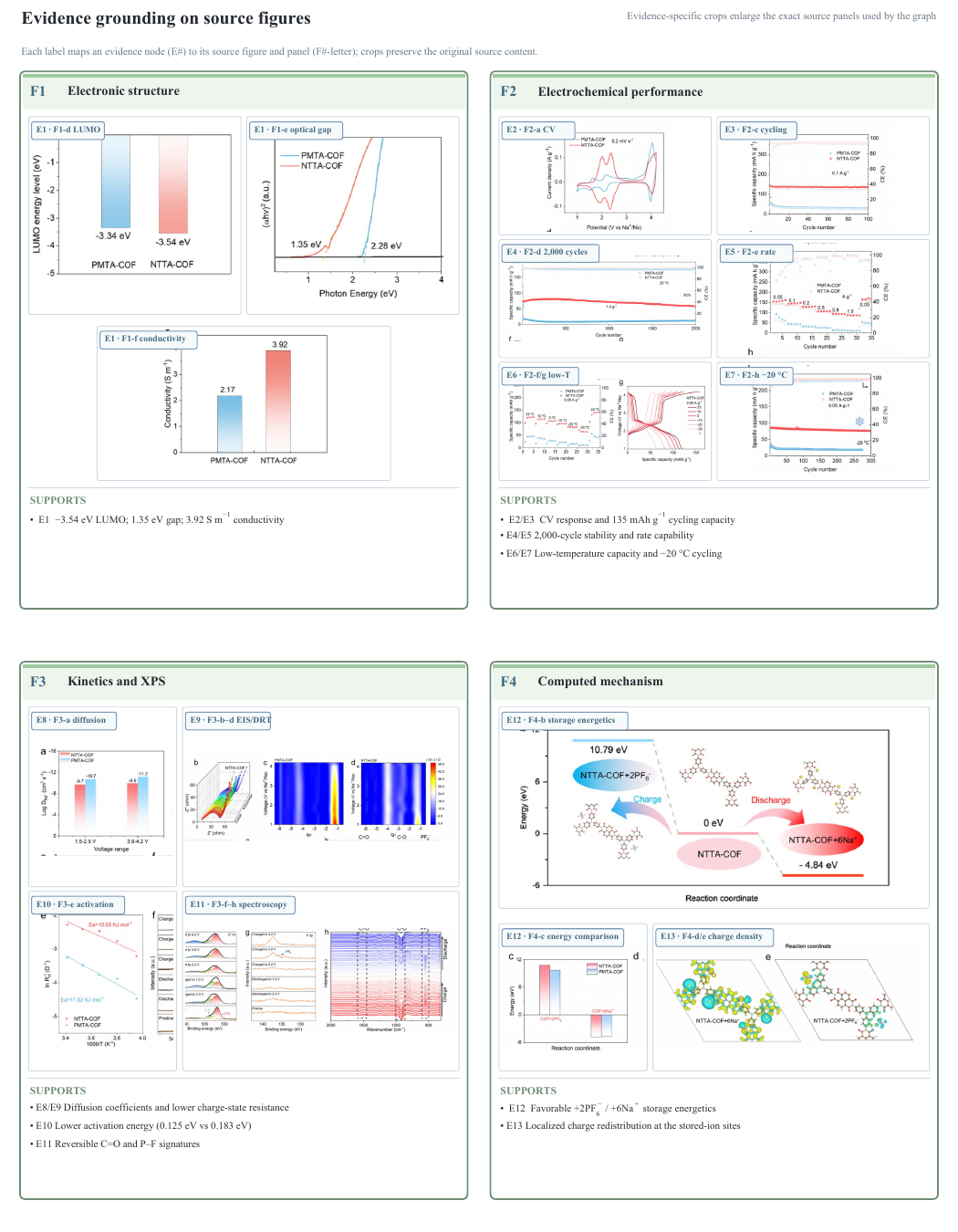}
    \caption{Evidence grounding for the audited argument graph. F1 grounds the electronic-structure evidence in E1; F2 grounds the room- and low-temperature performance evidence in E2--E7; F3 grounds the kinetic and spectroscopic evidence in E8--E11; and F4 grounds the computed storage mechanism in E12--E13. Each crop preserves the original source content and is labeled with the corresponding evidence node and source-panel location. Cropping is used only to improve legibility; source identity and panel-level traceability remain unchanged.}
    \label{fig:supp_evidence_grounding}
\end{figure*}

Figures~\ref{fig:supp_full_argument_graph} and~\ref{fig:supp_evidence_grounding} provide an audited example of the graph representation used by the benchmark. The example complements the aggregate distributions above with a concrete view of a higher-complexity graph and its source traceability. The complete graph contains 24 nodes and 32 directed relations at dependency depth five: one final claim and six intermediate claims are supported by 13 evidence nodes, each of which is grounded in one of four source figures. The claim hierarchy separates structural and electronic properties, electrochemical performance, and the proposed storage mechanism, while retaining the individual evidence values and source-panel identifiers.

Figure~\ref{fig:supp_evidence_grounding} expands the source side of the same graph. Instead of shrinking each composite source figure into a thumbnail, it shows evidence-specific crops labeled by both graph node and source panel. This view allows the numerical observations, curves, spectra, and computed structures to remain legible while preserving the E\#--F\# mapping used by the graph.

In the audited example, arrows are read from a conclusion to the evidence or lower-level claims that support it, and from an observation to the source panel in which it appears. For conceptual exposition, the main paper presents the same relations in the opposite inferential order---from visual evidence to claims. The two descriptions therefore encode the same dependency structure; only the direction in which the reader follows the arrows differs.

\subsection{Expert Verification Protocol}
\label{sec:supp_expert_verification}

Six experts verify the benchmark during construction. Four are discipline-matched reviewers with research backgrounds in biology, chemistry, computer science, or physics; two are senior reviewers with experience in scientific-figure interpretation, benchmark construction, or cross-disciplinary review. All have at least master's-level training in a relevant field. Every task is independently checked by one discipline-matched expert and one senior cross-reviewer. A second senior reviewer, who did not participate in the initial review of that task, adjudicates disagreements.

The review covers the question, supporting region boxes, evidence statements, claim and knowledge nodes, graph edges, reference final answer, overall answerability, and answer leakage. Discipline experts identify supporting regions and rewrite evidence as conservative propositions directly observable from the visual material. Cross-reviewers independently issue an accept-or-revise decision after inspecting the original visual, panel identity, and evidence coverage without seeing the first reviewer's decision. Region annotations are directly accepted when they refer to the same semantic region and reach IoU$\geq0.7$; otherwise a third reviewer adjudicates and, when needed, redraws the region. Evidence review checks entities, conditions, directions, comparisons, values, and units, and rejects unsupported inference or answer leakage. Graph review verifies node provenance and meaning, each support relation, connectivity and acyclicity, and the complete evidence $\rightarrow$ intermediate-claim $\rightarrow$ root-claim path. Initial, first-review, second-review, adjudication, and final-consensus versions are retained.

All 260 handoff tasks receive expert review, including 25 tasks that are rejected at this stage. Among the 235 retained tasks, 145 are directly accepted, 69 require minor revision, and 21 require major revision or redrawing; 24 tasks require third-reviewer adjudication. Ninety retained tasks require at least one object-level modification, while 145 pass without modification. Table~\ref{tab:supp_construction_review} reports the complete task-level disposition together with the object-level decisions; the adjudication column overlaps the disposition columns. Table~\ref{tab:supp_construction_agreement} reports agreement between the two initial reviewers before adjudication.

\begin{table*}[!t]
    \centering
    \small
    \setlength{\tabcolsep}{5pt}
    \begin{tabular}{lrrrrrr}
        \toprule
        \textbf{Object} & \textbf{Reviewed} & \textbf{Direct accept} & \textbf{Minor} & \textbf{Major/redraw} & \textbf{Rejected} & \textbf{Adjudicated} \\
        \midrule
        Benchmark tasks & 260 & 145 & 69 & 21 & 25 & 24 \\
        Questions & 235 & 180 & 45 & 10 & -- & 14 \\
        Region boxes & 2,107 & 1,790 & 242 & 75 & -- & 105 \\
        Evidence statements & 897 & 673 & 180 & 44 & -- & 81 \\
        Claim/knowledge nodes & 649 & 520 & 98 & 31 & -- & 52 \\
        Graph edges & 2,336 & 1,986 & 268 & 82 & -- & 128 \\
        Reference final answers & 235 & 211 & 19 & 5 & -- & 7 \\
        \bottomrule
    \end{tabular}
    \caption{Construction-stage expert review. The task row covers all 260 reviewed tasks: the 235 retained tasks are partitioned into direct acceptance, minor revision, and major revision/redrawing, while 25 are rejected. Rejected tasks are represented by their terminal task outcome; the remaining rows summarize final objects from retained tasks. Adjudication is an overlapping count.}
    \label{tab:supp_construction_review}
\end{table*}

\begin{table*}[!t]
    \centering
    \small
    \setlength{\tabcolsep}{8pt}
    \begin{tabular}{lrrr}
        \toprule
        \textbf{Decision unit} & \textbf{$n$} & \textbf{Exact agreement} & \textbf{Cohen's $\kappa$} \\
        \midrule
        Region accept/revise & 2,107 & 91\% & 0.80 \\
        Evidence-statement decision & 897 & 87\% & 0.77 \\
        Claim/knowledge-node decision & 649 & 87\% & 0.77 \\
        Graph-edge decision & 2,336 & 91\% & 0.83 \\
        Question accept/revise & 235 & 91\% & 0.83 \\
        Final-answer accept/revise & 235 & 96\% & 0.87 \\
        \bottomrule
    \end{tabular}
    \caption{Independent-reviewer agreement before construction-stage adjudication.}
    \label{tab:supp_construction_agreement}
\end{table*}

\section{Question Quality Control}
\label{sec:supp_question_quality_control}

The construction process begins with deterministic (rule-based) validation followed by an eight-dimensional language-model quality assessment. In the single candidate-task denominator used in the main text, deterministic validation removes 54\%, and the rubric stage filters a further 17\%. The depth check uses a minimum dependency depth of 3 together with graph-structure validity. Rubric failures, high-risk explicit checks, or failures on critical dimensions trigger conditional rewrites; 66 below-threshold flags, 140 rewrite triggers, and 130 accepted rewrites are overlapping event counts rather than a partition of tasks. After automated selection, 260 tasks enter expert review; 25 are rejected and 235 form the final benchmark. Table~\ref{tab:supp_task_flow} records the stage rates, terminal outcomes, and rewrite events.

The language-model assessment evaluates the generated question stem rather than a model response. Its purpose is to verify that a blind multi-hop question points to the intended visual materials, makes the evidence-to-claim path recoverable and genuinely necessary, states a clear scientific target, and does not reveal or presuppose the answer. Table~\ref{tab:supp_question_quality_rubric} gives the complete rubric in operational terms.

\begin{table*}[!t]
    \centering
    \small
    \setlength{\tabcolsep}{4pt}
    \renewcommand{\arraystretch}{1.16}
    \begin{tabular}{@{}p{0.19\textwidth}c p{0.37\textwidth}p{0.33\textwidth}@{}}
        \toprule
        \textbf{Dimension} & \textbf{Wt.} & \textbf{What is evaluated} & \textbf{Illustrative failure} \\
        \midrule
        \textbf{Anchor coverage}
        & 12
        & Whether the stem depends on the complete set of selected figures and tables without introducing unselected material.
        & A required figure is omitted, or the question asks the respondent to use an additional figure that is outside the task. \\

        \textbf{Anchor localization}
        & 8
        & Whether the wording lets the respondent reliably identify the intended figures, tables, or material range. A collective reference is acceptable when it is unambiguous.
        & The stem says only ``the figures'' even though the supplied materials contain several unrelated groups. \\

        \textbf{Evidence-path coverage}
        & 14
        & Whether the stem supplies enough neutral scientific context to recover the intended observation $\rightarrow$ interpretation $\rightarrow$ synthesis path, without spelling out the hidden reasoning steps.
        & ``What do these figures show?'' provides no indication of which objects, conditions, or relationship should organize the evidence. \\

        \textbf{Reasoning dependency}
        & 14
        & Whether answering requires combining local observations with at least one interpretive or synthesis step, rather than performing a single lookup or stating only a final conclusion.
        & One plotted value or one panel directly answers the question, so no cross-evidence inference is needed. \\

        \textbf{Target specificity}
        & 10
        & Whether the scientific object, condition, readout, comparison, or task context is precise enough to define the intended unknown without borrowing wording from the answer.
        & ``How well does the method work?'' leaves the dataset, metric, comparison, and operating condition unspecified. \\

        \textbf{Stem naturalness}
        & 22
        & Whether the stem is clear, readable, and phrased like a peer's scientific question rather than an exam prompt, procedural scaffold, checklist, or dense stack of technical nouns.
        & A long instruction enumerates every figure, metric, and reasoning step before asking for an ``overall conclusion.'' \\

        \textbf{Non-leakage}
        & 15
        & Whether the stem withholds result direction, rankings, trends, relation structure, intermediate conclusions, and the focal claim.
        & ``Why does method A outperform method B?'' discloses the comparison outcome that the respondent is meant to infer. \\

        \textbf{Answer shaping}
        & 5
        & Whether the stem avoids treating a target explanation or evaluative relation as already established and thereby forcing the response toward a predetermined conclusion.
        & ``How do these changes prove mechanism X?'' presupposes both the mechanism and the evidential relation to it. \\
        \bottomrule
    \end{tabular}
    \caption{Eight-dimensional quality rubric for generated blind multi-hop questions. The weights sum to 100 and produce a normalized overall score on a 0--100 scale. The last column illustrates common failure modes.}
    \label{tab:supp_question_quality_rubric}
\end{table*}

The overall passing threshold is 85. The three critical dimensions are anchor coverage, reasoning dependency, and non-leakage; a normalized score below 0.5 on any of them triggers revision even when the weighted total is near the threshold. Structural validation first checks dependency depth, graph validity, and the required structural elements. Question-level hard checks then flag high-risk defects that can be recognized directly, including exam-essay or proof-task phrasing, exposed solution procedures, over-generic stems, answer leakage, and answer-shaping formulations. A candidate is conditionally revised when a high-risk check fires, the language model recommends revision, the overall score is below 85, or a critical dimension falls below 0.5. Revision preserves the selected visual anchors, intended answer format, and multi-hop dependency while applying the smallest targeted wording change.

This construction-time rubric is distinct from the graph-aligned evaluation protocol below: it scores whether a question is suitable for inclusion, whereas Answer Accuracy, E-Cov., and C-Cov. score model responses to an accepted task.


\section{Evaluation Protocol}
\label{sec:supp_evaluation}

\subsection{Evaluation Settings}
\label{sec:supp_evaluation_settings}

We evaluate 235 questions with eight models under Caption-only, Direct Visual Reasoning, Evidence-Hint Reasoning, and Agentic Tool-Use, yielding 1,880 planned model--question observations per setting. Caption-only provides the question and complete source-paper figure/table captions, without image pixels, evidence hints, or paper body text. Direct Visual Reasoning provides the question and original visual references. Evidence-Hint Reasoning retains those inputs and supplies the annotated evidence text. Agentic Tool-Use preserves the original visual context and enables the model to crop and magnify selected regions. Answer accuracy always uses the planned denominator: missing evaluable answers count as incorrect. The numbers of available answers are 1,876/1,880 for Caption-only, 1,866/1,880 for Direct Visual Reasoning, 1,868/1,880 for Evidence-Hint Reasoning, and 1,873/1,880 for Agentic Tool-Use.

Difficulty is fixed before comparing models or evaluation settings. Each question receives a difficulty score from a weighted combination of graph-structural characteristics and visual-evidence demands, computed from the benchmark annotations rather than from model responses. The resulting strata contain 81 hard, 92 medium, and 62 easy questions, and the same fixed labels are used for every model and setting.

\subsection{Coverage Metrics}
\label{sec:supp_coverage_metrics}

Evidence coverage (E-Cov.) measures the fraction of required evidence nodes explicitly covered by the answer, and claim coverage (C-Cov.) analogously measures coverage of applicable intermediate-claim nodes. E-Cov. has 1,880 valid observations per setting. C-Cov. is defined for questions with a required intermediate claim and has 880 valid observations per setting. Together, the two metrics operationalize support-path reconstruction through observable answer text.

Answer correctness is judged through four binary atoms: whether the response gives a final conclusion, matches the gold final claim, preserves any applicable direction/polarity/comparison, and avoids a materially conflicting conclusion. The task-level Accuracy label is true only when all applicable atoms are labeled ``yes''; atoms marked ``not applicable'' are excluded, and a ``no'' on any applicable atom makes the response incorrect. Evidence and claim coverage are computed from independent binary node-match judgments. Missing model answers count as incorrect for Accuracy and as unmatched for applicable evidence and claim nodes.

\subsection{Human and Cross-Judge Audit}
\label{sec:supp_cross_judge_audit}

We conduct a human audit of 384 responses using a balanced stratified sample: four responses are sampled from each of the $8$ models $\times$ $4$ settings $\times$ $3$ fixed graph-structure difficulty strata. This $8\times4\times3\times4=384$ calculation is the denominator used in the audit tables. Four independent annotators with relevant master's-level scientific backgrounds participate, and each response is assigned to three annotators through a balanced rotation of the four possible three-person groups. A fifth, senior adjudicator with a doctorate and relevant research experience resolves every disputed atomic label after consulting the source paper and argument graph when necessary. Inter-annotator agreement is computed from the independent labels before adjudication; the adjudicator produces final human-consensus labels for comparison with Gemini 3.5 Flash.

Table~\ref{tab:supp_human_agreement} reports agreement at the correctness-atom and graph-node levels. Three-annotator unanimous agreement ranges from 86\% to 96\% for correctness atoms, is 83\% over 1,253 evidence-node decisions, and is 86\% over 273 intermediate-claim decisions. Against adjudicated human consensus, Gemini reaches 92\% exact agreement on the derived final Accuracy label, 85\% on evidence-node matches, and 87\% on intermediate-claim matches. Table~\ref{tab:supp_human_metric_validation} further shows close aggregate agreement for Accuracy, E-Cov., and C-Cov.

\begin{table}[!t]
    \centering
    \scriptsize
    \setlength{\tabcolsep}{3.5pt}
    \begin{tabular}{lrrrrr}
        \toprule
        \textbf{Metric} & \textbf{$n$} & \textbf{Human} & \textbf{Gemini} & \textbf{MAE} & \textbf{Spearman $\rho$} \\
        \midrule
        Accuracy & 384 & 52\% & 52\% & -- & -- \\
        E-Cov. & 384 & 54\% & 54\% & 0.08 & 0.85 \\
        C-Cov. & 192 & 45\% & 45\% & 0.10 & 0.82 \\
        \bottomrule
    \end{tabular}
    \caption{Metric-level validation against adjudicated human consensus. Scores are means over the audited responses.}
    \label{tab:supp_human_metric_validation}
\end{table}

\begin{table*}[!t]
    \centering
    \scriptsize
    \setlength{\tabcolsep}{4pt}
    \begin{tabular}{lrrr@{\hspace{7mm}}rrr}
        \toprule
        & \multicolumn{3}{c}{\textbf{Independent human annotators}}
        & \multicolumn{3}{c}{\textbf{Human consensus vs.\ Gemini}} \\
        \cmidrule(lr){2-4}\cmidrule(lr){5-7}
        \textbf{Decision unit} & \textbf{$n$} & \textbf{Exact} & \textbf{Fleiss' $\kappa$}
        & \textbf{$n$} & \textbf{Exact} & \textbf{Cohen's $\kappa$} \\
        \midrule
        Gives final conclusion & 384 & 96\% & 0.92 & 384 & 96\% & 0.81 \\
        Matches gold final claim & 384 & 91\% & 0.80 & 384 & 91\% & 0.84 \\
        Preserves direction/polarity/comparison & 200 & 86\% & 0.81 & 200 & 91\% & 0.82 \\
        No materially conflicting conclusion & 384 & 91\% & 0.82 & 384 & 92\% & 0.81 \\
        Evidence-node match & 1,253 & 83\% & 0.77 & 1,253 & 85\% & 0.80 \\
        Intermediate-claim-node match & 273 & 86\% & 0.76 & 273 & 87\% & 0.82 \\
        Derived final Accuracy label & -- & -- & -- & 384 & 92\% & 0.80 \\
        \bottomrule
    \end{tabular}
    \caption{Human inter-annotator agreement before adjudication and Gemini 3.5 Flash agreement with adjudicated human consensus. Human exact agreement requires all three assigned annotators to agree.}
    \label{tab:supp_human_agreement}
\end{table*}

\begin{figure*}[!t]
    \centering
    \includegraphics[width=\textwidth]{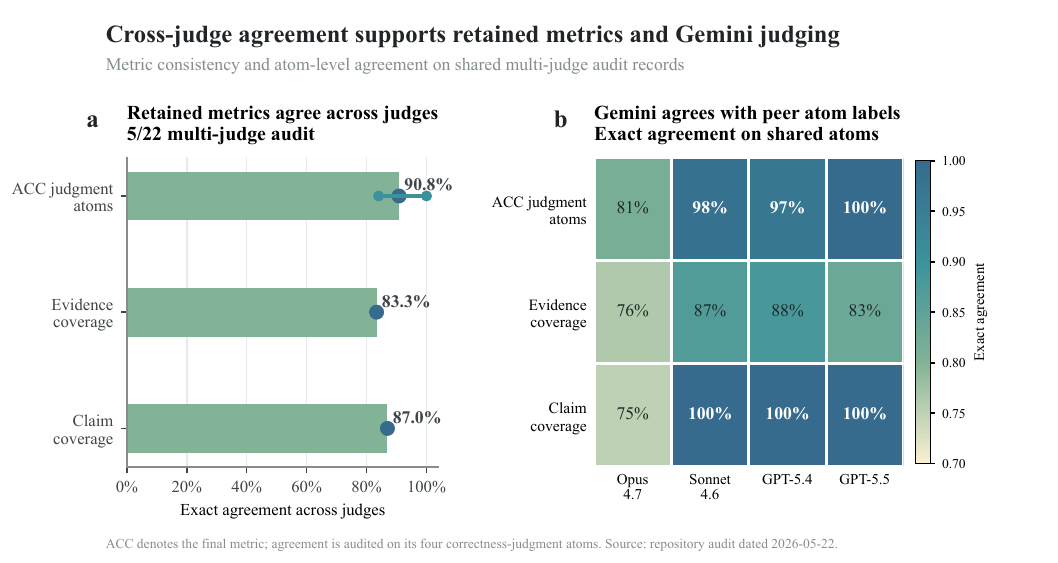}
    \caption{Cross-judge consistency. (a) Exact agreement for correctness, E-Cov., and C-Cov. (b) Pairwise exact agreement between Gemini 3.5 Flash and four peer judges.}
    \label{fig:supp_judge_consistency}
\end{figure*}

Figure~\ref{fig:supp_judge_consistency} summarizes a multi-judge audit on shared records. Mean exact agreement over the four answer-correctness atoms is 90.8\% (atom range 84.0--100.0\%; 25--29 comparable groups). Exact agreement is 83.3\% for evidence coverage over 120 groups and 87.0\% for claim coverage over 23 groups. In the pairwise audit, Gemini 3.5 Flash agrees with the four peer judges on 81.4--100.0\% of shared correctness atoms ($n=24$--60) and 76.3--88.2\% of shared evidence-node labels ($n=18$--69). The claim-node estimates use 3--12 shared labels, with fewer than 10 for three of the four peers. Accordingly, the cross-judge consistency conclusion is anchored in the more extensively supported correctness and evidence-node comparisons.

\section{Baseline Reasoning and Evidence Recovery}
\label{sec:supp_rq1}

\subsection{Caption-Only Diagnostic}
\label{sec:supp_caption_only}

Caption-only measures how much of each task can be solved from source-paper captions without visual pixels. Table~\ref{tab:supp_caption_only} reports the complete model-level results.

    \begingroup
    \scriptsize
    \setlength{\tabcolsep}{2pt}%
    \renewcommand{\arraystretch}{1.02}%
    \input{tables/supp_caption_only_diagnostics}    \endgroup


Pooled Caption-only accuracy is 38.7\%, below the 46.1\% achieved under Direct Visual Reasoning, with E-Cov. and C-Cov. of 26.5\% and 26.8\%. Accuracy decreases for six models; MiniMax-M3 gains 0.5 points and Intern-S2 gains 9.8 points. Captions therefore provide useful semantic cues for some tasks but do not replace the original visuals overall.

\subsection{Discipline and Difficulty Strata}
\label{sec:supp_rq1_strata}

Table~\ref{tab:supp_rq1_stratified} reports the complete Direct Visual Reasoning breakdown by discipline and overall; the main results table gives the complementary fixed graph-annotation strata.

\input{tables/supp_rq1_stratified_accuracy}

Pooled accuracy is highest for chemistry (56.6\%) and lowest for physics (41.6\%) in this analysis set. The discipline comparison is descriptive; the main results table reports the corresponding fixed difficulty-stratum comparison.

\subsection{Coverage and Correctness}
\label{sec:supp_rq1_coverage_correctness}

\begin{figure}[!htbp]
    \centering
    \includegraphics[width=\columnwidth]{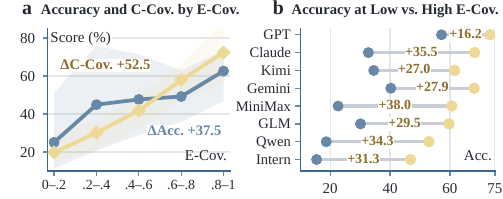}
    \caption{Coverage diagnostics under Direct Visual Reasoning. (a) Pooled accuracy and C-Cov. by E-Cov. bin, with model ranges shaded. (b) Per-model accuracy at low ($<0.3$) and high ($\geq0.8$) E-Cov.}
    \label{fig:reasoning_recovery_diagnostics}
\end{figure}

Figure~\ref{fig:reasoning_recovery_diagnostics} summarizes the relationship between coverage and correctness under Direct Visual Reasoning. Panel (a) reports pooled accuracy and C-Cov. across E-Cov. bins, together with the range of model-level accuracy. Panel (b) compares each model's accuracy at low E-Cov. ($<0.3$) and high E-Cov. ($\geq0.8$).

Table~\ref{tab:supp_rq1_recovery} provides a stricter lower-tail check, defining low E-Cov. as below 0.2 while retaining high E-Cov. at 0.8 or above. The localization and error-routing analyses use 0.3 as their declared low-coverage threshold.

    \begingroup
    \scriptsize
    \setlength{\tabcolsep}{2pt}%
    \renewcommand{\arraystretch}{1.02}%
    \input{tables/supp_rq1_recovery_threshold}    \endgroup

The low--high comparison holds within every model. Pooled accuracy is 25.1\% among the 486 observations with E-Cov.$<0.2$, compared with 62.6\% among the 554 observations with E-Cov.$\geq0.8$, a 37.5-point difference. This descriptive comparison establishes a consistent association between output evidence coverage and correctness; the following intervention analyses then examine performance under direct evidence access.

MiniMax-M3 illustrates why coverage complements accuracy. Under Direct Visual Reasoning it has the highest model-level E-Cov. (61.5\%) and C-Cov. (51.7\%), together with 45.5\% accuracy. Coverage records whether annotated evidence and intermediate claims appear in the output, while accuracy captures whether that information is integrated and mapped to the correct final option. The MiniMax-M3 result therefore identifies a model whose outputs frequently express the intended support chain while achieving moderate final-answer accuracy.

\section{Evidence-Access Interventions}
\label{sec:supp_interventions}
\label{sec:supp_rq2}

\subsection{Evidence Hint Gains}
\label{sec:supp_rq2_gains}

Table~\ref{tab:supp_rq2_hint} gives model-level overall results for Evidence-Hint Reasoning, together with paired rescue and loss counts; the main results table gives the complementary fixed graph-annotation strata.

    \begingroup
    \scriptsize
    \setlength{\tabcolsep}{2pt}%
    \renewcommand{\arraystretch}{1.02}%
    \input{tables/supp_rq2_hint_diagnostics}    \endgroup

Evidence-Hint Reasoning improves pooled accuracy and C-Cov. across the benchmark, while the main table gives the corresponding fixed-stratum view. C-Cov. increases for all eight models, with model-level changes between +15.8 and +33.1 points.

Paired transitions separate gross accuracy changes from individual outcome reversals. Across the 1,880 planned pairs, Evidence-Hint Reasoning rescues 597 initially wrong Direct Visual Reasoning answers and loses 82 initially correct answers, for a net gain of 515 correct observations (+27.4 points). Every model has substantially more rescues than losses. These gains quantify the combined effect of the evidence-explicit intervention, including greater evidence salience and reduced visual transcription and disambiguation burden.

\subsection{Agentic Tool-Use Changes}
\label{sec:supp_rq3}

\paragraph{Model-level changes.}
\label{sec:supp_rq3_model_changes}

Table~\ref{tab:supp_rq3_crop} compares Agentic Tool-Use with Direct Visual Reasoning for accuracy, E-Cov., C-Cov., and paired answer transitions.

    \begingroup
    \scriptsize
    \setlength{\tabcolsep}{2pt}%
    \renewcommand{\arraystretch}{1.02}%
    \input{tables/supp_rq3_crop_diagnostics}    \endgroup


Agentic Tool-Use produces a smaller and less uniform intervention than Evidence Hint. It yields 260 rescues and 175 losses, for a net gain of 85 correct observations (+4.5 points). Seven models improve both E-Cov. and C-Cov., but Intern-S2 is the exception: its accuracy rises from 26.8\% to 34.0\% while E-Cov. falls from 41.8\% to 32.6\% and C-Cov. falls from 31.2\% to 22.1\%.

Accuracy and coverage summarize different properties of the 235 outputs. Accuracy counts correct final answers, whereas E-Cov. and C-Cov. average the fractions of annotated nodes expressed in each response. Because these metrics aggregate separate response properties, their changes can be distributed differently across question pairs. Agentic Tool-Use can therefore increase the total number of correct answers while reducing the average amount of annotated support stated in the outputs. For Intern-S2, the supported conclusion is an output-level divergence: higher final-answer accuracy accompanies lower explicit evidence and claim coverage.

\subsection{Localization and Observed Rescue}
\label{sec:supp_rq3_localization}

Table~\ref{tab:supp_rq3_hits} focuses on 299 observations that are initially wrong, have Direct Visual Reasoning E-Cov.$<0.3$, and invoke the crop tool.

\input{tables/supp_rq3_hit_count}


The answer-rescue rate increases from 20.0\% with no annotated-region hit to 38.7\% with at least four hits. The shares of runs with increased E-Cov. are 48.6\%, 46.4\%, 55.7\%, and 64.5\% across the four hit bins, respectively, with the two highest values in the upper-hit bins. Together, these descriptive results associate greater annotated-region access with stronger evidence recovery and answer rescue.

A key-region hit requires a crop-tool box and gold region on the same reference to reach IoU$\geq0.5$. This definition makes hit count a conservative geometric measure: crops can contain useful legends, axes, captions, neighboring structure, or semantically relevant but loosely aligned regions below the strict threshold. Hit count therefore records geometric access, while E-Cov. records the evidence expressed from that access.

\section{Error Attribution and Qualitative Analysis}
\label{sec:supp_error_attribution}

\subsection{Routing Rules}
\label{sec:supp_error_routing_rules}

The analysis unit is one paired model--question observation among the 1,014 incorrect Direct Visual Reasoning answers. Let $E_D$ and $C_D$ denote its Direct Visual Reasoning E-Cov. and C-Cov., and let $H$ and $K$ denote final-answer correctness under Evidence-Hint Reasoning and Agentic Tool-Use. A coverage value is low below 0.3 and high at or above 0.8.

The Direct Visual Reasoning group $G$ is assigned as follows. It is \emph{claim/interpretation insufficient} when $E_D\geq0.8$ and $C_D$ is either inapplicable or below 0.3. It is \emph{final answer still wrong} when $E_D\geq0.8$ and applicable $C_D\geq0.8$. All remaining cases, including observations with intermediate E-Cov. or C-Cov., are assigned to \emph{evidence insufficient}.

Localization under Agentic Tool-Use is \emph{no call}, \emph{no hit}, \emph{partial hit}, or \emph{all hit}; a hit requires same-reference IoU$\geq0.5$, and all hit requires every annotated key region to be hit. For a wrong response with high E-Cov., we use the final-answer-residual label only when applicable C-Cov. is also high; otherwise, including when C-Cov. is inapplicable, we use the interpretation-residual label. The label ``Agentic Tool-Use rescue, no qualifying hit'' records either the absence of a crop meeting this annotated-region hit criterion or the absence of an observed crop call; output outcomes separately capture useful information available outside strict overlap.

We apply the following ordered rules so that the eight labels are mutually exclusive and exhaustive. First, we inspect the Direct Visual Reasoning response. If its E-Cov. is at least 0.8 while C-Cov. is either inapplicable or below 0.3, we label the case ``claim/interpretation insufficient.'' If both E-Cov. and applicable C-Cov. are at least 0.8, we label it ``final answer still wrong.'' All other Direct Visual Reasoning errors are classified as ``evidence insufficient.'' This preserves the earliest informative trace before using outcomes from later settings.

For the remaining cases, we compare correctness under Evidence-Hint Reasoning and Agentic Tool-Use. When the hint response is wrong but the tool-use response is correct, the case is a ``cross-setting Agentic Tool-Use rescue.'' When both are wrong, a hint E-Cov. below 0.8 is labeled ``Evidence Hint not fully reflected''; otherwise, high evidence coverage is separated into an interpretation residual or a final-answer residual according to claim coverage. When both intervention responses are correct, a tool run with no call or no qualifying region hit is recorded as an ``Agentic Tool-Use rescue, no qualifying hit,'' while a run with a qualifying hit is assigned to access/localization. Finally, when the hint response is correct but the tool-use response is wrong, incomplete localization is assigned to access/localization; after all annotated regions are hit, Tool-Use E-Cov. below 0.3 is ``visual extraction: low,'' E-Cov. from 0.3 to below 0.8 is ``visual extraction: partial,'' and higher E-Cov. is separated into an interpretation or final-answer residual by claim coverage.


\subsection{Model-Level Subtype Composition}
\label{sec:supp_error_routing_models}

\begin{figure*}[!t]
    \centering
    \includegraphics[width=\textwidth]{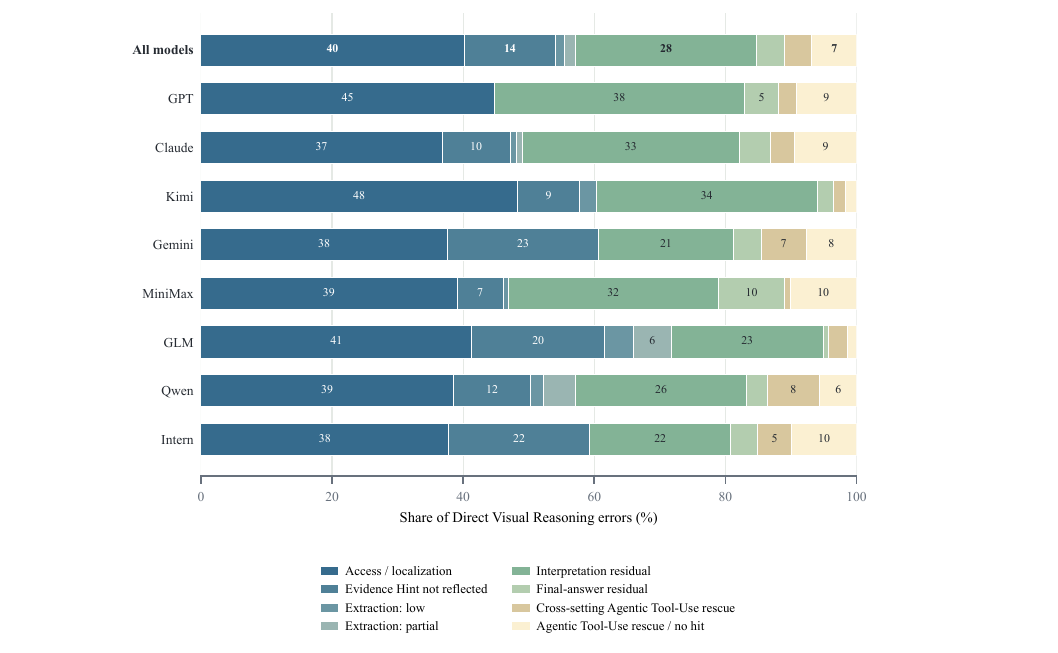}
    \caption{Per-model composition of the eight mutually exclusive error subtypes at IoU 0.5. Each horizontal row is normalized over that model's incorrect Direct Visual Reasoning outputs, while the accompanying count gives its error denominator. Colors retain the same subtype mapping across models; subtype assignment follows the ordered routing rules described in the text.}
    \label{fig:supp_error_routing_models}
\end{figure*}

Figure~\ref{fig:supp_error_routing_models} shows that access/localization is the largest individual subtype for every model, while the balance between Evidence Hint uptake, downstream interpretation, and audit categories varies. The overall partition contains 407 access/localization errors, 142 Evidence-Hint-not-reflected errors, 14 low-extraction errors, 17 partial-extraction errors, 279 interpretation residuals, 43 final-answer residuals, 43 cross-setting Agentic Tool-Use rescues, and 69 Agentic Tool-Use rescues without a qualifying hit. These sum exactly to 1,014.

The main text reports six broader subtypes by combining the two extraction labels and the two audit labels. The resulting counts are 407 access/localization errors (40.1\%), 142 Evidence-Hint-not-reflected (gold-underuse) errors (14.0\%), 31 visual-extraction errors (3.1\%), 279 interpretation residuals (27.5\%), 43 final-answer residuals (4.2\%), and 112 audit/other cases (11.0\%). Thus, the three evidence-related groups total 57.2\% of errors, the two downstream-reasoning groups total 31.8\%, and the audit/other group accounts for the remaining 11.0\%. The eight-way breakdown below retains the finer distinctions used for sensitivity analysis and representative cases.

\subsection{Interpreting Zero-Count Subtypes}
\label{sec:supp_error_routing_zero_count_subtypes}

Zero entries in Figure~\ref{fig:supp_error_routing_models} report observed support under the router's joint criteria. The two visual-extraction labels activate when Evidence-Hint Reasoning is correct, Agentic Tool-Use is wrong, and all annotated regions are hit, so their branch-specific support is substantially smaller than the 76--172 Direct Visual Reasoning errors contributed by each model. GPT-5.5's Hint-not-reflected observations provide a complementary example: its ten cases with both Evidence-Hint Reasoning and Agentic Tool-Use wrong all have Hint E-Cov. of at least 0.8 and are assigned to downstream residuals. The zero entries therefore summarize empirical subtype prevalence under the declared routing rules and branch eligibility.

\subsection{Threshold Sensitivity}
\label{sec:supp_error_routing_threshold_sensitivity}

The error-routing sensitivity analysis reruns the fine-grained router at IoU thresholds of 0.3 and 0.7, bracketing the default threshold of 0.5. Table~\ref{tab:supp_error_routing_sensitivity} reports aggregate subtype counts, while Table~\ref{tab:supp_error_routing_transitions} lists the nonzero label transitions between adjacent thresholds. Of the 1,014 routed observations, 54 labels change from IoU 0.3 to 0.5 and 37 change from 0.5 to 0.7. The access/localization subtype remains the largest at every threshold (405, 407, and 400 observations, respectively), and the two downstream subtypes together remain stable (325, 322, and 319). Stricter overlap primarily moves successful Agentic Tool-Use cases into the no-qualifying-hit audit category and moves complete-localization extraction cases back into access/localization.

    \begingroup
    \scriptsize
    \setlength{\tabcolsep}{2pt}%
    \renewcommand{\arraystretch}{1.02}%
    \input{tables/supp_rq4_threshold_sensitivity}    \endgroup


\subsection{Representative Error-Pattern Cards}
\label{sec:supp_error_routing_cases}

\begin{figure*}[!p]
    \centering
    \includegraphics[width=0.62\textwidth]{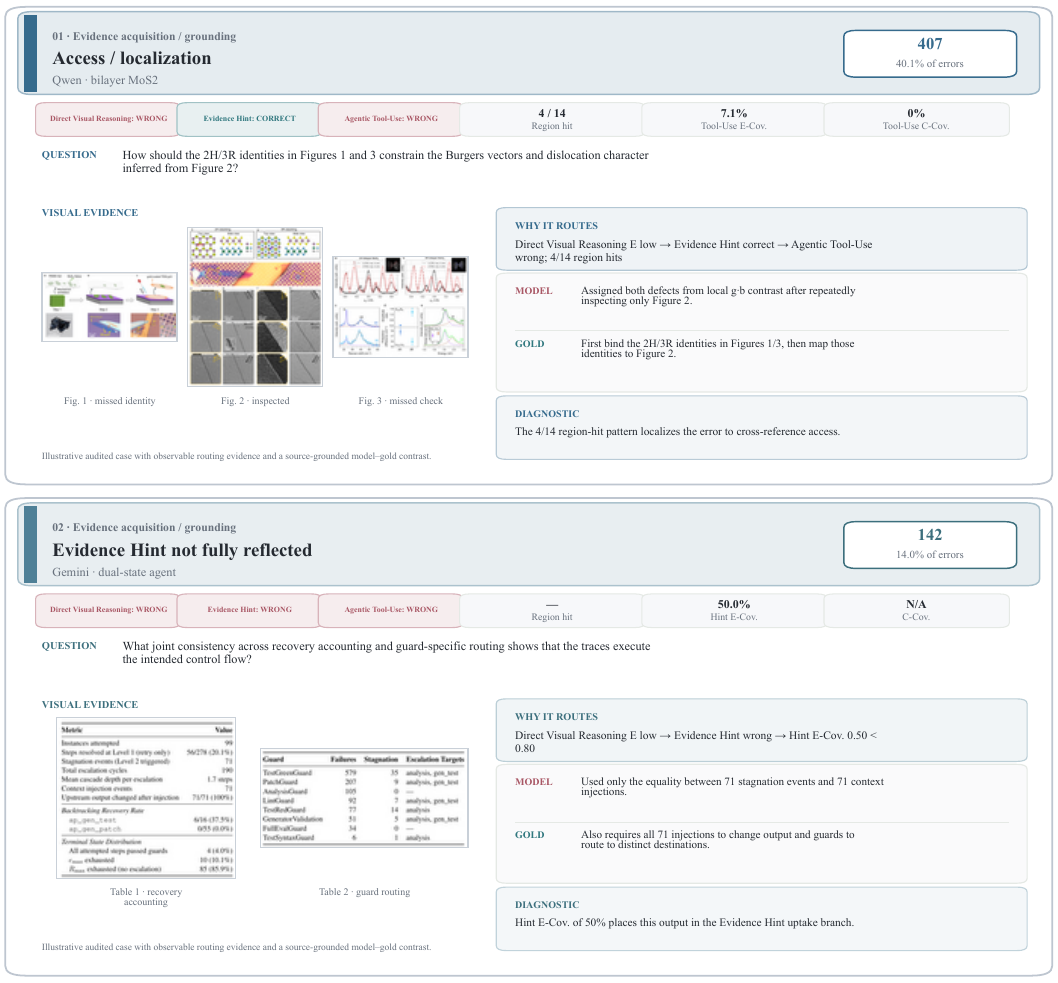}
    \caption{Audited evidence-acquisition examples. The upper card (Case 1) is routed to access/localization because Agentic Tool-Use reaches 4 of 14 annotated regions and omits two cross-reference checks. The lower card (Case 2) is routed to Evidence Hint not fully reflected because its answer covers part of the supplied constraints (Hint E-Cov. 50.0\%). Each card reports the cross-setting outcomes, routing measurements, inspected visual evidence, model--gold contrast, and diagnostic implication.}
    \label{fig:supp_cases_01_02}
    \vspace{6pt}
    \includegraphics[width=0.62\textwidth]{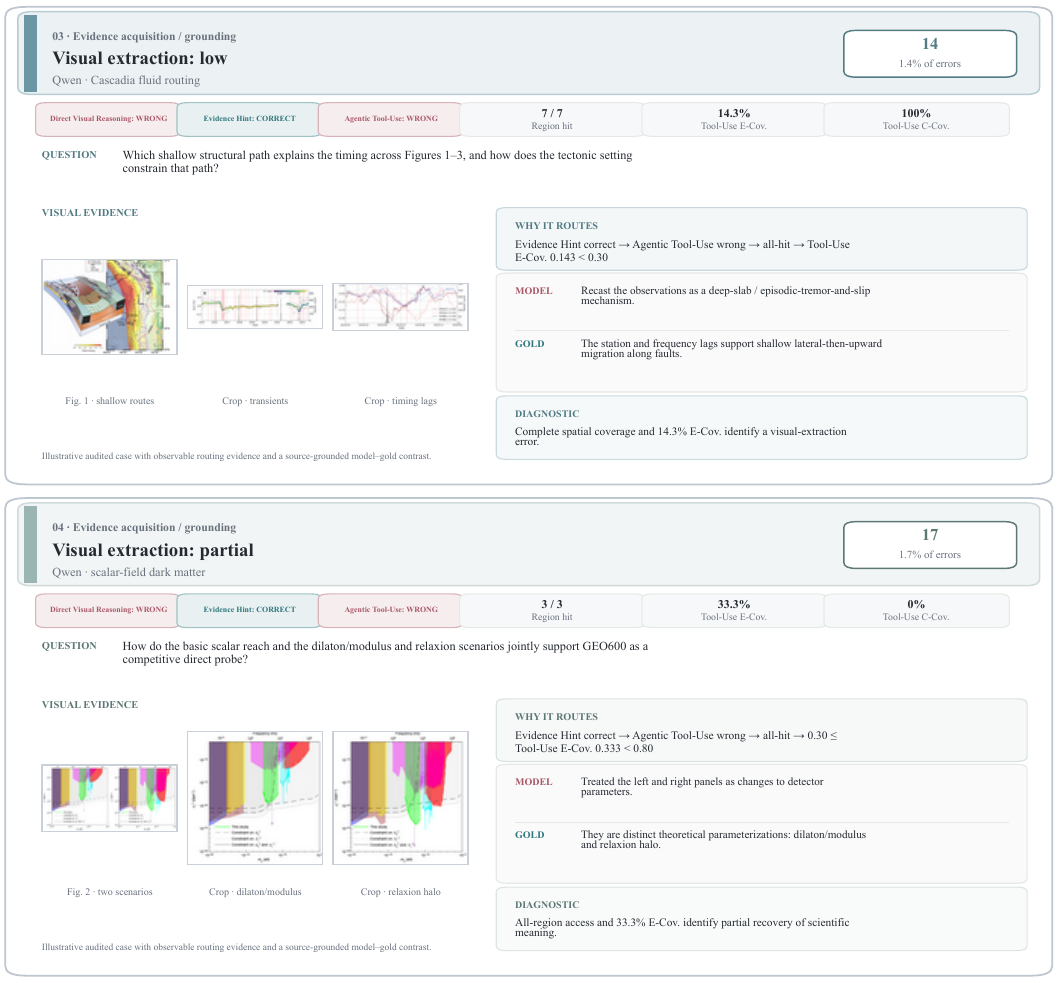}
    \caption{Audited visual-extraction examples after complete annotated-region access. The upper card (Case 3) has 7/7 region hits and 14.3\% Tool-Use E-Cov., yielding the low-extraction label. The lower card (Case 4) has 3/3 hits and 33.3\% Tool-Use E-Cov., yielding the partial-extraction label. The model--gold contrasts distinguish complete geometric access from recovery of scientific meaning.}
    \label{fig:supp_cases_03_04}
\end{figure*}

\begin{figure*}[!p]
    \centering
    \includegraphics[width=0.62\textwidth]{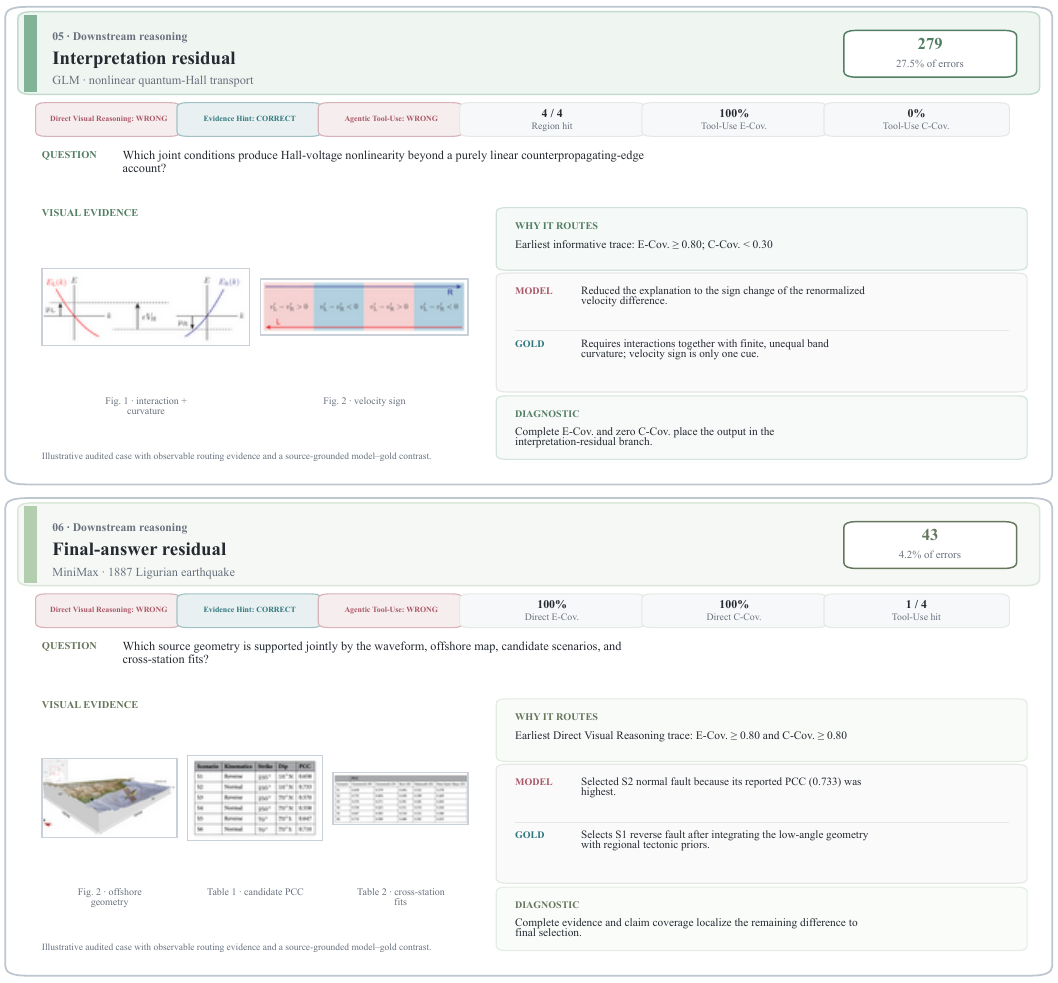}
    \caption{Audited downstream-reasoning examples. The upper card (Case 5) has 100\% Tool-Use E-Cov. and 0\% Tool-Use C-Cov., placing it in the interpretation-residual category. The lower card (Case 6) has 100\% Direct Visual Reasoning E-Cov. and C-Cov. and selects a different final fault plane, placing it in the final-answer-residual category.}
    \label{fig:supp_cases_05_06}
    \vspace{6pt}
    \includegraphics[width=0.62\textwidth]{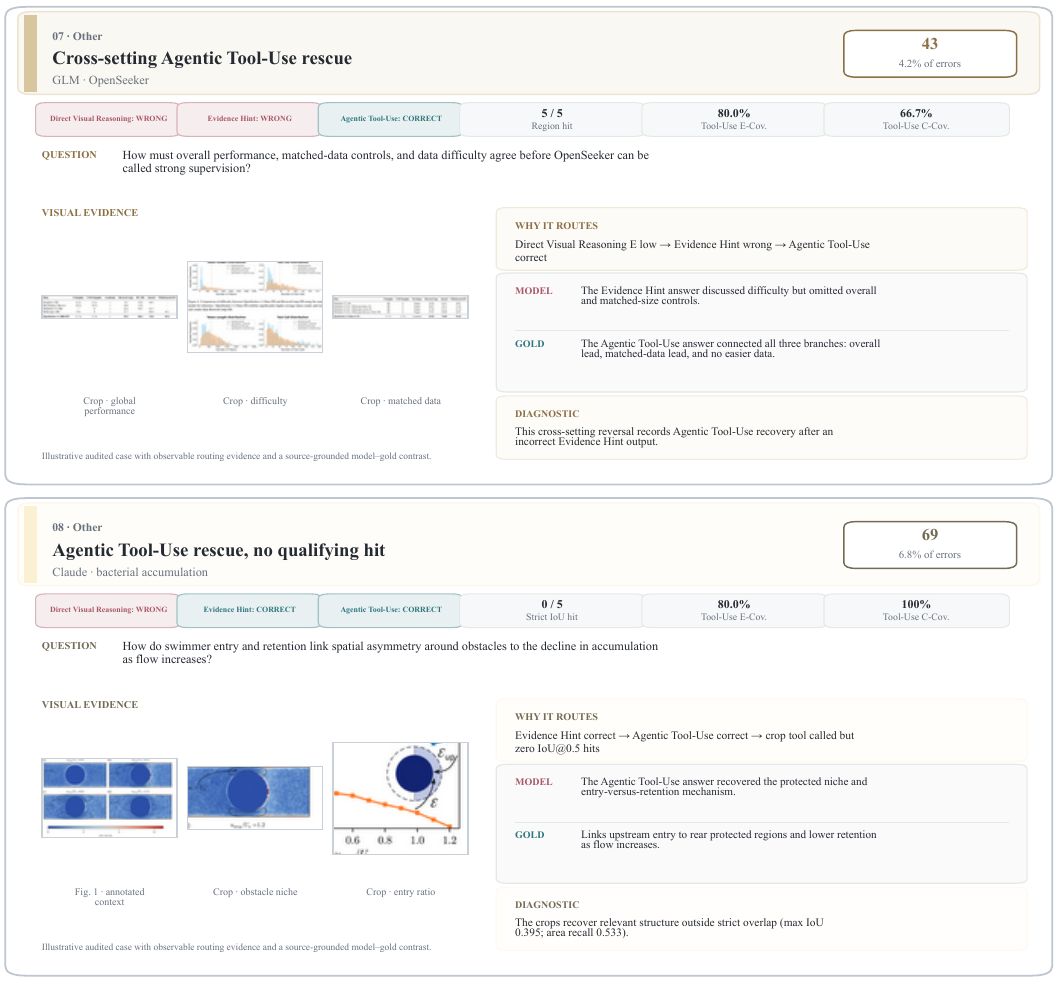}
    \caption{Audited intervention and metric-boundary examples. The upper card (Case 7) is a cross-setting reversal: Agentic Tool-Use is correct after covering the three comparison branches following an incorrect Evidence Hint output. The lower card (Case 8) is an Agentic Tool-Use rescue with zero strict IoU@0.5 hits and semantically relevant structure in the crops. These labels characterize observed cross-setting reversals and geometric-proxy boundaries.}
    \label{fig:supp_cases_07_08}
\end{figure*}

Figures~\ref{fig:supp_cases_01_02}, \ref{fig:supp_cases_03_04}, \ref{fig:supp_cases_05_06}, and~\ref{fig:supp_cases_07_08} present one audited illustrative case for each subtype. Every card follows the same reading order: cross-setting correctness and coverage/localization values, the source or actual crop views, the model--gold contrast, and the routing implication. The subtype counts and shares quantify prevalence over all 1,014 Direct Visual Reasoning errors, and each selected example provides a concrete instance of its routing rule.

\paragraph{Access and Evidence Hint uptake.}

Figure~\ref{fig:supp_cases_01_02} contrasts two evidence-acquisition patterns. In Case 1, Agentic Tool-Use inspects 4 of 14 annotated regions and misses the cross-reference identities needed to interpret the central panel. In Case 2, the supplied Evidence Hint is partially reflected (E-Cov. 50.0\%): the answer uses one equality while omitting the output-change and guard-routing constraints. The pair separates incomplete visual access from incomplete uptake of already supplied evidence.

\paragraph{Extraction after complete localization.}

Figure~\ref{fig:supp_cases_03_04} restricts attention to two examples with the same operational localization status: both hit every annotated region, Evidence Hint answers correctly, and Agentic Tool-Use remains wrong. Case 3 covers 14.3\% of the required evidence and substitutes a different geophysical mechanism, whereas Case 4 reaches 33.3\% and extracts part of the scientific role of the two scalar-field scenarios. The contrast operationalizes the low and partial visual-extraction branches and separates geometric access from evidence recovery.

\paragraph{Downstream interpretation and final selection.}

Figure~\ref{fig:supp_cases_05_06} distinguishes two errors in outputs that explicitly cover all annotated evidence. Case 5 has complete Tool-Use E-Cov. but zero C-Cov.; the answer notices a velocity cue without forming the required joint claim about interactions and curvature. Case 6 already covers both the required evidence and intermediate claims under Direct Visual Reasoning, yet selects the wrong fault plane during final aggregation. The pair therefore separates missing claim expression from an incorrect final-answer selection in the observed outputs.

\paragraph{Cross-setting and localization-proxy audits.}

Figure~\ref{fig:supp_cases_07_08} records two audit categories at the boundaries of the intervention and localization metrics. Case 7 is a cross-setting reversal in which Agentic Tool-Use combines all three required comparison branches and answers correctly after an incorrect Evidence Hint output. Case 8 answers correctly with zero strict IoU@0.5 hits; its crops contain semantically useful obstacle structure, with maximum IoU 0.395 and area recall 0.533. Together, the cases preserve the distinction between Evidence Hint outcomes and oracle behavior, and between strict region overlap and useful visual access.

\input{tables/supp_prompt_cards}

%% file: tables/supp_caption_only_diagnostics.tex
\begin{table*}[!t]
\centering
\scriptsize
\setlength{\tabcolsep}{7pt}
\begin{tabular}{lrrrrr}
\toprule
Model & Answered & Acc. & $\Delta$ Acc. & E-Cov. & C-Cov. \\
\midrule
GPT-5.5 & 235/235 & 49.4 & -18.3 & 28.7 & 31.6 \\
Claude Opus 4.8 & 231/235 & 42.1 & -12.8 & 27.8 & 29.0 \\
Kimi K2.7 & 235/235 & 48.5 & -2.1 & 31.1 & 33.3 \\
Gemini 3.1 Pro & 235/235 & 25.1 & -25.1 & 18.6 & 16.4 \\
MiniMax-M3 & 235/235 & 46.0 & +0.5 & 34.5 & 38.1 \\
GLM-5V Turbo & 235/235 & 37.0 & -4.3 & 27.2 & 25.2 \\
Qwen3.7 Plus & 235/235 & 25.1 & -6.4 & 20.3 & 16.1 \\
Intern-S2 & 235/235 & 36.6 & +9.8 & 24.2 & 24.8 \\
\midrule
\textbf{Pooled} & 1876/1880 & 38.7 & -7.4 & 26.5 & 26.8 \\
\bottomrule
\end{tabular}
\caption{Caption-only diagnostic results. Inputs contain the question and complete source-paper figure/table captions, without image pixels, evidence hints, or paper body text. Accuracy changes are relative to Direct Visual Reasoning.}
\label{tab:supp_caption_only}
\end{table*}

%% file: tables/supp_rq1_stratified_accuracy.tex
\begin{table*}[!t]
\centering
\scriptsize
\setlength{\tabcolsep}{5.0pt}
\begin{tabular}{lrrrrr}
\toprule
Model & Bio. & Chem. & Comp. & Phys. & Overall \\
\midrule
GPT-5.5 & 54.0 & 76.5 & 73.0 & 68.2 & 67.7 \\
Claude Opus 4.8 & 48.0 & 76.5 & 61.9 & 45.5 & 54.9 \\
Kimi K2.7 & 46.0 & 55.9 & 58.7 & 45.5 & 50.6 \\
Gemini 3.1 Pro & 46.0 & 61.8 & 49.2 & 48.9 & 50.2 \\
MiniMax-M3 & 46.0 & 58.8 & 46.0 & 39.8 & 45.5 \\
GLM-5V Turbo & 46.0 & 50.0 & 36.5 & 38.6 & 41.3 \\
Qwen3.7 Plus & 40.0 & 41.2 & 33.3 & 21.6 & 31.5 \\
Intern-S2 & 30.0 & 32.4 & 23.8 & 25.0 & 26.8 \\
\midrule
\textbf{Pooled} & 44.5 & 56.6 & 47.8 & 41.6 & 46.1 \\
\bottomrule
\end{tabular}
\caption{Direct Visual Reasoning accuracy (\%) by scientific discipline and overall. Difficulty-stratified headline results are reported in the main results table using fixed graph-annotation strata.}
\label{tab:supp_rq1_stratified}
\end{table*}

%% file: tables/supp_rq1_recovery_threshold.tex
\begin{table*}[!t]
\centering
\scriptsize
\setlength{\tabcolsep}{7pt}
\begin{tabular}{lrrrrr}
\toprule
Model & Low $n$ & Low Acc. & High $n$ & High Acc. & $\Delta$ \\
\midrule
GPT-5.5 & 47 & 51.1 & 90 & 73.3 & +22.3 \\
Claude Opus 4.8 & 46 & 28.3 & 85 & 68.2 & +40.0 \\
Kimi K2.7 & 72 & 31.9 & 78 & 61.5 & +29.6 \\
Gemini 3.1 Pro & 68 & 27.9 & 47 & 68.1 & +40.1 \\
MiniMax-M3 & 38 & 15.8 & 99 & 60.6 & +44.8 \\
GLM-5V Turbo & 63 & 25.4 & 57 & 59.6 & +34.3 \\
Qwen3.7 Plus & 73 & 16.4 & 51 & 52.9 & +36.5 \\
Intern-S2 & 79 & 11.4 & 47 & 46.8 & +35.4 \\
\midrule
\textbf{Pooled} & 486 & 25.1 & 554 & 62.6 & +37.5 \\
\bottomrule
\end{tabular}
\caption{Within-model Direct Visual Reasoning accuracy at low E-Cov. ($<0.2$) and high E-Cov. ($\geq0.8$); $\Delta$ is high minus low in percentage points.}
\label{tab:supp_rq1_recovery}
\end{table*}

%% file: tables/supp_rq2_hint_diagnostics.tex
\begin{table*}[!t]
\centering
\scriptsize
\setlength{\tabcolsep}{6pt}
\begin{tabular}{lrrrrr}
\toprule
Model & Direct Acc. & Hint Acc. ($\Delta$) & $\Delta$C-Cov. & Rescue & Loss \\
\midrule
GPT-5.5 & 67.7 & 86.8 (+19.2) & +33.1 & 53 & 8 \\
Claude Opus 4.8 & 54.9 & 78.3 (+23.4) & +16.5 & 60 & 5 \\
Kimi K2.7 & 50.6 & 76.6 (+26.0) & +26.5 & 74 & 13 \\
Gemini 3.1 Pro & 50.2 & 70.3 (+20.0) & +15.8 & 59 & 12 \\
MiniMax-M3 & 45.5 & 75.3 (+29.8) & +20.5 & 84 & 14 \\
GLM-5V Turbo & 41.3 & 71.5 (+30.2) & +20.8 & 85 & 14 \\
Qwen3.7 Plus & 31.5 & 68.5 (+37.0) & +27.3 & 92 & 5 \\
Intern-S2 & 26.8 & 60.4 (+33.6) & +22.5 & 90 & 11 \\
\midrule
\textbf{Pooled} & 46.1 & 73.5 (+27.4) & +22.9 & 597 & 82 \\
\bottomrule
\end{tabular}
\caption{Overall Evidence-Hint Reasoning diagnostics by model; parentheses show accuracy changes from Direct Visual Reasoning. Difficulty-stratified results are reported in the main results table.}
\label{tab:supp_rq2_hint}
\end{table*}

%% file: tables/supp_rq3_crop_diagnostics.tex
\begin{table*}[!t]
\centering
\scriptsize
\setlength{\tabcolsep}{7pt}
\begin{tabular}{lrrrrr}
\toprule
Model & Tool-Use Acc. ($\Delta$) & Tool-Use E-Cov. ($\Delta$) & Tool-Use C-Cov. ($\Delta$) & Rescue & Loss \\
\midrule
GPT-5.5 & 69.8 (+2.1) & 63.7 (+5.2) & 53.4 (+7.3) & 20 & 15 \\
Claude Opus 4.8 & 60.0 (+5.1) & 67.9 (+8.6) & 54.0 (+2.5) & 30 & 18 \\
Kimi K2.7 & 52.8 (+2.2) & 55.6 (+4.7) & 43.5 (+6.8) & 29 & 24 \\
Gemini 3.1 Pro & 52.3 (+2.1) & 56.2 (+12.3) & 50.5 (+11.1) & 23 & 18 \\
MiniMax-M3 & 50.6 (+5.1) & 67.5 (+6.0) & 54.1 (+2.4) & 37 & 25 \\
GLM-5V Turbo & 44.3 (+3.0) & 57.4 (+9.8) & 44.3 (+10.4) & 37 & 30 \\
Qwen3.7 Plus & 40.9 (+9.4) & 54.8 (+9.7) & 40.6 (+6.0) & 47 & 25 \\
Intern-S2 & 34.0 (+7.2) & 32.6 (-9.2) & 22.1 (-9.0) & 37 & 20 \\
\midrule
\textbf{Pooled} & 50.6 (+4.5) & 57.0 (+5.9) & 45.3 (+4.7) & 260 & 175 \\
\bottomrule
\end{tabular}
\caption{Agentic Tool-Use diagnostics by model; parentheses show changes from Direct Visual Reasoning.}
\label{tab:supp_rq3_crop}
\end{table*}

%% file: tables/supp_rq3_hit_count.tex
\begin{table}[t]
\centering
\scriptsize
\setlength{\tabcolsep}{4.5pt}
\begin{tabular}{lrrr}
\toprule
Hits & $n$ & E-Cov. improved & Answer rescued \\
\midrule
0 & 105 & 51 (48.6\%) & 21 (20.0\%) \\
1 & 84 & 39 (46.4\%) & 22 (26.2\%) \\
2-3 & 79 & 44 (55.7\%) & 23 (29.1\%) \\
$\geq 4$ & 31 & 20 (64.5\%) & 12 (38.7\%) \\
\bottomrule
\end{tabular}
\caption{Agentic Tool-Use outcomes by annotated-region hit count for initially incorrect, low-coverage observations. A hit requires same-reference IoU$\geq0.5$.}
\label{tab:supp_rq3_hits}
\end{table}

%% file: tables/supp_rq4_threshold_sensitivity.tex
\begin{table*}[!t]
\centering
\scriptsize
\setlength{\tabcolsep}{7pt}
\begin{tabular}{lrrr}
\toprule
Error subtype & IoU 0.3 & IoU 0.5 & IoU 0.7 \\
\midrule
Access / localization & 405 & 407 & 400 \\
Evidence Hint not fully reflected & 142 & 142 & 142 \\
Visual extraction: low & 26 & 14 & 7 \\
Visual extraction: partial & 30 & 17 & 12 \\
Interpretation residual & 281 & 279 & 276 \\
Final-answer residual & 44 & 43 & 43 \\
Cross-setting rescue & 43 & 43 & 43 \\
Rescue / no qualifying hit & 43 & 69 & 91 \\
\midrule
\textbf{Total} & \textbf{1,014} & \textbf{1,014} & \textbf{1,014} \\
\bottomrule
\end{tabular}
\caption{Error-routing sensitivity analysis: error-subtype counts under alternative same-reference IoU thresholds. Every column contains the same 1,014 initially incorrect Direct Visual Reasoning observations.}
\label{tab:supp_error_routing_sensitivity}
\end{table*}

\begin{table*}[!t]
\centering
\scriptsize
\setlength{\tabcolsep}{5pt}
\begin{tabular}{lllr}
\toprule
Threshold change & Previous label & New label & $n$ \\
\midrule
0.3 $\rightarrow$ 0.5 & Access / localization & Rescue / no qualifying hit & 26 \\
0.3 $\rightarrow$ 0.5 & Visual extraction: partial & Access / localization & 13 \\
0.3 $\rightarrow$ 0.5 & Visual extraction: low & Access / localization & 12 \\
0.3 $\rightarrow$ 0.5 & Interpretation residual & Access / localization & 2 \\
0.3 $\rightarrow$ 0.5 & Final-answer residual & Access / localization & 1 \\
\midrule
0.5 $\rightarrow$ 0.7 & Access / localization & Rescue / no qualifying hit & 22 \\
0.5 $\rightarrow$ 0.7 & Visual extraction: low & Access / localization & 7 \\
0.5 $\rightarrow$ 0.7 & Visual extraction: partial & Access / localization & 5 \\
0.5 $\rightarrow$ 0.7 & Interpretation residual & Access / localization & 3 \\
\midrule
\multicolumn{3}{r}{\textbf{Changed at 0.3 $\rightarrow$ 0.5}} & \textbf{54} \\
\multicolumn{3}{r}{\textbf{Changed at 0.5 $\rightarrow$ 0.7}} & \textbf{37} \\
\bottomrule
\end{tabular}
\caption{Error-routing label transitions between adjacent IoU thresholds; unlisted observations are unchanged.}
\label{tab:supp_error_routing_transitions}
\end{table*}

%% file: tables/supp_prompt_cards.tex
\clearpage
\raggedbottom
\section{Prompt Catalogue}
\label{sec:supp_prompt_catalogue}

The catalogue reports the task-defining instructions, dynamic input fields, and downstream-consumed output contracts used by the benchmark. Repeated discipline-specific wording, long pedagogical examples, expanded sample payloads, and serialization-equivalent historical protocol variants are summarized rather than reproduced four times. Black-tabbed boxes are native searchable LaTeX listings; long boxes continue automatically across columns or pages.

\subsection{Benchmark Construction}
\subsubsection{Argument Graph and Visual Grounding}

\begin{PromptCard}{C-G1  Argument-Graph Construction}
[SCOPE]
n=235; shared by all four prompt profiles; source: figure_qa/prompts/*/graph.py

[SYSTEM PROMPT]
You are an expert scientific reader. Build an argument graph for ONE paper.

## Node roles
- **claim**: what the authors conclude or argue.
- **evidence**: one ref-grounded empirical or textual observation *from this paper* that is used to support a claim. Make the `label` self-contained and specific enough to stand on its own. It does **not** need to be limited to one sentence: use one to a few sentences when needed to preserve the assay, comparison, direction, and key quantitative detail, but keep one node focused on one observation rather than mixing unrelated findings.
- **claim** labels should also be informative rather than compressed. They do **not** need to be limited to one sentence: use one to a few sentences when needed to state the conclusion precisely, but keep one node focused on one claim rather than bundling several independent conclusions together.
- **figure_ref**: one cited paper figure used as a proof anchor. Treat multiple sub-panels from the same overall figure as the same `figure_ref` unless the paper clearly cites them as separate proof anchors.
- **table_ref**: one cited paper table used as a proof anchor. Treat one overall table as one `table_ref` unless the paper clearly separates table sections into distinct proof anchors.
- **established_basis**: non-figure proof the paper relies on as **widely accepted** (textbook fact, standard method, prior consensus, or clearly framed "it is known that ..."). Use a short `label` quoting or paraphrasing what is taken for granted. Use `attributes.citation` when the text cites references for that basis.

## Article-source attribution
- Every `claim` and `evidence` node must contain a non-empty `source_attributions` array that identifies where its information appears in the supplied paper text.
- Each source attribution must contain `section_title` (the nearest section/subsection heading, or null when unavailable) and `source_excerpt` (a short verbatim passage sufficient to verify the node label).
- Use multiple source-attribution items only when a node genuinely synthesizes non-contiguous passages. Keep each excerpt minimal and exact; do not paraphrase, invent wording, or cite a passage that merely discusses a related topic.
- `figure_ref`, `table_ref`, and `established_basis` nodes use an empty `source_attributions` array because their provenance is represented by their citation/label contract.

## Mandatory proof for every evidence
- **Every `evidence` node MUST have `proven_by` anchors, including exactly one outgoing edge to either `figure_ref` or `table_ref`.**
- An `evidence` node may also have additional outgoing `proven_by` edges to `established_basis` when background knowledge is needed to interpret that same ref-grounded observation.
- Direction: `source` = evidence id, `target` = proof id, `type` = `proven_by`.
- Claims are linked by `supported_by`: **source** is always the `claim` being supported; **target** is the supporting node, usually `evidence`, but **may also be another `claim` or an `established_basis`** when the paper chains conclusions or explicitly invokes accepted prior knowledge. **Do not** force everything into a single-hop triple when the narrative is multi-step.
- `figure_ref` and `table_ref` nodes are proof anchors only: they may be the target of `proven_by`, but they must never be the target of `supported_by`.

## Chain arguments (multi-hop is allowed; mildly prefer mixed support when clearly warranted)
- If the text naturally presents stepwise reasoning (broader conclusion -> sub-result / premise), **slightly prefer** preserving the intermediate `claim` node(s) and multiple `supported_by` edges instead of collapsing everything into one hop.
- When a higher-level claim is justified by both (a) direct observations and (b) an intermediate conclusion, represent both supports explicitly (`claim_high -> evidence` and `claim_high -> claim_mid`).
- Do not over-fragment: if the paper states a direct support relation without a meaningful intermediate step, keep it single-hop.
- Typical path shapes include `claim_main ->supported_by-> claim_sub ->supported_by-> evidence ->proven_by-> figure_ref/table_ref`, or `claim_high ->supported_by-> evidence` plus `claim_high ->supported_by-> claim_mid`.
- A `claim` that only serves as a step may have **both** outgoing `supported_by` (to evidence and/or sub-claims) and incoming `supported_by` (from a broader claim).

## established_basis usage
- If the paper cites prior work as the reason an observation holds, model that observation as `evidence` and add `proven_by` to both the single `figure_ref`/`table_ref` anchor and the needed `established_basis` node(s).
- If the paper directly invokes accepted prior knowledge as part of a higher-level inference, you may link `claim ->supported_by-> established_basis`.
- Do not output floating `established_basis` nodes. Every basis must be used by at least one `evidence` via `proven_by` or by at least one `claim` via `supported_by`.
- Experimental setup, assay condition, grouping scheme, or measurement protocol from this paper is **not** `established_basis`; do not invent them as separate basis nodes.

## First-round ref policy
- This is the first round. Do not try to bind refs to real image files or final reference ids.
- Use temporary ids such as `graph_ref_1`, `graph_ref_2`, `graph_ref_3`.
- Each `figure_ref` / `table_ref` label must be detailed enough for later semantic matching: state what the figure/table is about, the main objects or assays, and the key comparison/setup dimension when available.
- Keep `attributes.citation` when the text names a figure/table citation such as `Fig. 2B` or `Table 3`.
- One ref may support multiple distinct evidence nodes if the paper reads multiple separate observations from the same overall figure/table.
- Do not split one overall figure into multiple ref nodes just because different panels are mentioned, unless the text clearly treats them as separate proof anchors.

## Scale
- Expect **more nodes** when many figures are discussed; **roughly 20--80 nodes** is normal for figure-heavy papers. Preserve distinct proof anchors, but do not create extra ref nodes for every panel crop.

## Output
- **Exactly one** JSON object, no markdown fences, no commentary.
- List **every** `supported_by` and `proven_by` in the top-level **`edges` array only**. Do **not** nest `proven_by` (or any edges) inside `nodes`.
- Unique `id` per node (ASCII snake_case or short alphanumeric).
- Use English `label` when the paper is English; otherwise keep the original language.
- Include every listed JSON key.
- `attributes` must always be an object with key `citation`; use `null` when no in-text citation is available.

Allowed node `type` values:
  claim, evidence, figure_ref, table_ref, established_basis

Allowed edge `type` values:
  supported_by, proven_by

## Minimal sufficient support (per claim and per evidence)
- For each **claim** `C`, treat its outgoing `supported_by` edges as a **conjunction** of premises: list only targets the text actually uses together to establish `C`. **Avoid** parallel edges that duplicate the same reasoning; prefer distinct, non-overlapping premises.
- For each **evidence** `E`, each `proven_by` target must anchor what the sentence genuinely uses as proof in the paper.
- Each evidence must have exactly one ref anchor (`figure_ref` or `table_ref`). If the paper combines two refs, split that into multiple evidence nodes plus an intermediate claim when needed.
- One ref may be reused by multiple evidence nodes when the paper draws multiple distinct observations from the same figure or table.
- Do not attach figure/table/basis nodes that are not substantively used for that observation.
- Evidence nodes should be concise but not underspecified: include the measured object, contrast/group, and direction of effect when the text gives them. Multi-sentence labels are acceptable when needed for fidelity.
- When quantitative detail is explicit and central (counts, percentages, fold-changes, significance, named genes/methods), prefer keeping it in the node `label` and/or `attributes` rather than omitting it.

`figure_ref` / `table_ref` fields for this round: only output `id`, `type`, `label`, and `attributes.citation`. Use temporary ids and keep the label detailed enough for later matching.
`established_basis` fields: `label` (required), `attributes.citation` nullable for in-text ref tokens.

[USER TEMPLATE -- DYNAMIC FIELDS]
Article id: <ARTICLE_ID>
Figure/table candidate list: <REFERENCE_INVENTORY>
Main paper text: <ARTICLE_TEXT>

[OUTPUT CONTRACT]
{
  "nodes": [{
    "id": "<UNIQUE_ID>",
    "type": "<claim|evidence|figure_ref|table_ref|established_basis>",
    "label": "<SELF_CONTAINED_LABEL>",
    "attributes": {"citation": "<CITATION_OR_NULL>"},
    "source_attributions": [{
      "section_title": "<NEAREST_HEADING_OR_NULL>",
      "source_excerpt": "<SHORT_VERBATIM_PASSAGE>"
    }]
  }],
  "edges": [{"source":"<ID>","target":"<ID>","type":"<supported_by|proven_by>"}]
}
Claim/evidence nodes require source_attributions. Reference and established-basis nodes use an empty source_attributions array.
\end{PromptCard}
\begin{PromptCard}{C-G2/3  Reference Linking and Conditional Visual Review}
[SCOPE]
C-G2 n=235; C-G3 conditional n=68

[C-G2 -- BATCH TEXT LINKING]
You batch-link graph-side figure/table references to real layout candidates for one paper.

Task:
- You will receive all graph-side refs from the extracted argument graph.
- You will also receive all real figure/table candidates extracted from the paper layout.
- For each graph ref, choose the single best matching candidate id, or null when text alone is not sufficient.

Rules:
- Match by caption semantics first.
- `graph_ref_type` and `ref_type` must agree.
- Use citation strings such as `Fig. 2B` or `Table 3` only as weak hints, never as the primary key.
- Do not match a main-text figure to an `Extended Data Figure`, or the reverse, unless the citation family explicitly agrees.
- Prefer candidates whose caption meaning matches both the graph ref label and its neighboring evidence/claim context.
- When text-only matching is genuinely uncertain, set `needs_visual_disambiguation=true` and provide `alternative_candidate_ids`.
- Do not force a choice when multiple candidates remain plausible from text alone.

Return exactly one JSON object with field:
- `matches`: array of objects containing
  - `graph_ref_id`
  - `matched_candidate_id`
  - `confidence`
  - `alternative_candidate_ids`
  - `needs_visual_disambiguation`

Final output contract:
{"matches":[{"graph_ref_id":"<input graph ref id>","matched_candidate_id":"<best candidate id or null>","confidence":"<high|medium|low|null>","alternative_candidate_ids":["<candidate id>"],"needs_visual_disambiguation":false}]}

[INPUT / OUTPUT]
Input: graph_refs[] plus extracted figure/table candidates[] and neighboring evidence/claim labels.
Output: {"matches":[{
  "graph_ref_id":"<INPUT_ID>",
  "matched_candidate_id":"<BEST_ID_OR_NULL>",
  "confidence":"<high|medium|low|null>",
  "alternative_candidate_ids":["<ID>"],
  "needs_visual_disambiguation":false
}]}

[C-G3 -- CONDITIONAL VISUAL REVIEW]
You visually disambiguate one graph-side figure/table reference against a short candidate list.

Task:
- Read the graph ref label, its citation hint, and its neighboring evidence/claim labels.
- Compare them against the provided candidate captions.
- Inspect the attached candidate preview images in the same order as `review_candidates`.
- Return the single best-matching candidate id, or null when the match remains genuinely uncertain.

Rules:
- Match graph semantics to both caption meaning and visual layout/content.
- Use the attached images only to break ambiguities that remained after text matching.
- Treat `Figure N` and `Extended Data Figure N` as different figure families unless the citation family explicitly matches.
- Do not force a choice if the candidates still cannot be distinguished confidently.

Return exactly one JSON object with:
- `graph_ref_id`
- `matched_candidate_id`

Final output contract:
{"graph_ref_id":"<input graph ref id>","matched_candidate_id":"<best candidate id or null>"}

[INPUT / OUTPUT]
Input: one unresolved graph_ref, a short candidate list, and candidate preview images.
Output: {"graph_ref_id":"<INPUT_ID>","matched_candidate_id":"<BEST_ID_OR_NULL>"}
\end{PromptCard}
\begin{PromptCard}{C-G4  Visual-Evidence Rewrite}
[SCOPE]
shared contract; BIO 50, CHEM 34, CS 63, PHY 88

[SYSTEM PROMPT]
You rewrite paper-text evidence into figure/table-observable evidence for a figure-grounded benchmark.

You will receive one resolved reference and the evidence nodes currently linked to it. Inspect the attached reference image/table preview.

Task for each evidence item:
1. Confirm the best reference id from the provided reference context. In this first implementation there is usually one resolved ref; still return `selected_ref_id`.
2. Rewrite the original evidence into `visual_evidence_text`: a statement that can be directly observed from this figure/table using visible panels, axes, legends, labels, annotations, table values, or caption-defined setup.
3. Remove mechanisms, causal explanations, author conclusions, clinical/biological interpretations, and paper-text-only claims that are not directly visible in the selected reference.
4. If only part of the evidence is visible, keep only the directly visible part.
5. If no directly observable support is found, leave `visual_evidence_text` empty or extremely conservative.

Rules:
- Do not invent numeric values, groups, markers, genes, panels, trends, or labels not visible in the image/table or caption-defined setup.
- Caption/setup terms may be used to name conditions or measurements, but caption conclusions must not be copied as observations unless they are visibly supported.
- Prefer concrete visual language: higher/lower, increase/decrease, overlap, enrichment, localization, panel/axis/legend/table-value references.
- Avoid words that overclaim beyond the reference: proves, demonstrates mechanism, causes, confirms, therapeutic potential, due to, therefore.
- Return exactly one JSON object with field `items` and one item for every input evidence id.

[USER TEMPLATE -- DYNAMIC FIELDS]
reference: id, type, label, citation, caption, and non-result context
evidence_items[]: evidence_id, original_evidence_text, parent_claim_labels, candidate_ref_ids, and article_source_attributions
attachment: <RESOLVED_FIGURE_OR_TABLE_IMAGE>

[OUTPUT CONTRACT]
{"items":[{
  "evidence_id":"<INPUT_EVIDENCE_ID>",
  "selected_ref_id":"<REFERENCE_ID_OR_NULL>",
  "visual_evidence_text":"<DIRECTLY_OBSERVABLE_EVIDENCE_OR_EMPTY>"
}]}

[DISCIPLINE PROFILE ADAPTATION]
The task and output contract are shared across all four profiles. Only terminology and examples change:
- Biology: biological objects, compartments, conditions, assays, and response patterns.
- Chemistry: compounds, materials, reaction conditions, spectra, morphology, and readouts.
- Computer Science: methods, datasets, metrics, benchmarks, ablations, and failure modes.
- Physics: physical systems, observables, parameters, regimes, spectra, and measurement channels.
\end{PromptCard}
\subsubsection{Question and Gold-Answer Construction}

\begin{PromptCard}{C-Q1  Evidence-Dialogue Planning}
[SCOPE]
stage observed in n=214 benchmark samples

[SYSTEM PROMPT]
Write diverse observation questions from the given image/table evidence and one evidence statement.

You will receive:
- article background;
- the canonical program-side data for one evidence step;
- the image/table directly tied to that evidence step.

Task: use the current image/table, the current evidence statement, and the abstract task family to propose multiple observation questions about directly readable phenomena. These questions should target the evidence encoded by the current statement, but they must ask about observable facts rather than restating the conclusion itself.

Requirements:
- Focus on what should be observed, compared, or measured from the current material; do not restate a higher-level conclusion.
- Prefer quantitative comparisons when the material supports them: who is higher, by how much, how much earlier, which row/column is larger, whether a curve collapses or rebounds, and similar directly readable facts.
- The question may mention figure/table/panel/location cues when useful.
- If a figure/table is mentioned, use the question-time alias labels supplied by the user JSON; figure and table numbering restart from 1 inside the current QA task.
- Keep every sample local, observable, and non-conclusive; do not reveal the final claim in advance.
- Return all requested samples in one JSON object.
- Every sample must target the same evidence step, but the samples should be meaningfully different in angle or emphasis rather than trivial paraphrases.
- Do not ask for the already-compressed conclusion as the direct answer.
- Return one shared `key_region` field for the whole evidence step, not separate boxes per sample.

Return one JSON object only, with no markdown. Fields:
- `step`
- `derived_node_id`
- `key_region`
- `samples`, each containing only `observation_question`

[USER TEMPLATE -- DYNAMIC FIELDS]
article background; question-time reference aliases; one canonical evidence step; article source attributions; requested sample count; attached figure/table

[OUTPUT CONTRACT]
{
  "step":"<OBSERVATION_STEP>",
  "derived_node_id":"<EVIDENCE_NODE_ID>",
  "key_region":[x_min,y_min,x_max,y_max],
  "samples":[{"observation_question":"<LOCAL_VISUAL_QUESTION>"}]
}

[DISCIPLINE PROFILE ADAPTATION]
The task and output contract are shared across all four profiles. Only terminology and examples change:
- Biology: biological objects, compartments, conditions, assays, and response patterns.
- Chemistry: compounds, materials, reaction conditions, spectra, morphology, and readouts.
- Computer Science: methods, datasets, metrics, benchmarks, ablations, and failure modes.
- Physics: physical systems, observables, parameters, regimes, spectra, and measurement channels.
\end{PromptCard}
\begin{PromptCard}{C-Q2  Transition-Bridge Planning}
[SCOPE]
stage observed in n=87 benchmark samples

[SYSTEM PROMPT]
Write one bridge question that links local observations to the next supported conclusion.

You will receive:
- article background;
- the final preferred claim;
- the canonical program-side data for one non-final claim step;
- the observation questions already written for the supporting evidence steps of this claim step.

Task: write one bridge question that asks what interpretation or local conclusion is supported when those observation questions are answered together, without directly stating the current claim statement or the final preferred claim..

Requirements:
- `bridge_question` must be a real inferential question, not a phrase stub.
- It must clearly depend on the supplied observation questions.
- Professional terms from the paper are allowed when they help precision.
- If a figure/table is mentioned, use the question-time alias labels supplied by the user JSON; figure and table numbering restart from 1 inside the current QA task.
- Compare the supplied observations against the current claim statement and the available `source_basis_ids` / `available_basis_contexts`.
- Do not directly restate the current claim statement.
- Do not directly restate the final preferred claim.

Return one JSON object only, with no markdown. Fields:
- `step`
- `derived_node_id`
- `bridge_question`

[USER TEMPLATE -- DYNAMIC FIELDS]
article background; preferred final claim; one intermediate-claim step; supporting observation questions; available established-basis context; claim article source attributions

[OUTPUT CONTRACT]
{
  "step":"<INTERMEDIATE_CLAIM_STEP>",
  "derived_node_id":"<CLAIM_NODE_ID>",
  "bridge_question":"<OBSERVATION_TO_INTERPRETATION_QUESTION>"
}

[DISCIPLINE PROFILE ADAPTATION]
The task and output contract are shared across all four profiles. Only terminology and examples change:
- Biology: biological objects, compartments, conditions, assays, and response patterns.
- Chemistry: compounds, materials, reaction conditions, spectra, morphology, and readouts.
- Computer Science: methods, datasets, metrics, benchmarks, ablations, and failure modes.
- Physics: physical systems, observables, parameters, regimes, spectra, and measurement channels.
\end{PromptCard}
\begin{PromptCard}{C-Q3  Blind-Question Generation}
[SCOPE]
stage observed in n=235 benchmark samples

[SYSTEM PROMPT]
Write a blind question from slim material hints.

The user JSON will provide article background, selected anchors, final_claim_text, neutral observation focuses, optional bridge questions, bridge mode, and a minimal output contract. Use those inputs to write only the question stem. Do not output the answer, explanation, or reasoning chain.

Use final_claim_text as the guarded target conclusion: it is provided only to prevent the question from drifting away from the intended final claim. The stem must make the respondent inspect the selected figures/tables, establish local observations, and reach that target through a recoverable observation-to-interpretation path grounded in the supplied observation focuses and bridge questions. Those inputs are latent guidance only: the stem does not need to spell out the bridge structure in advance. Do not turn final_claim_text into the stem or into answer-guiding shells such as "what overall conclusion follows", "which explanation is best supported", "which interpretation is most consistent with", or "what, if anything, does ...".

Task: write one difficult blind question. The ideal respondent cannot see the article text and only has access to the relevant figures/tables, so they must inspect the materials, establish local observations, connect them to at least one local inference, and then synthesize the overall answer.

Question-design requirements:
- Write one clear biology research question that sounds like a peer's real scientific question, not a synthesis scaffold or a material-by-material checklist.
- Keep one main scientific unknown in view. You may name one or two high-level biological objects, response contexts, or readout families when needed, but do not recover specificity by stacking categories.
- Prefer direct question forms that name the main biological unknown, relationship, or consistency condition to be resolved. If the question stays recoverable without opening on figure/table labels, prefer object-first, mechanism-first, or consistency-check phrasing over anchor-first openings such as "Across Figures ...", "In Figure ...", or "From Figures ...". Evidence-to-interpretation wording is fine when it stays natural, but do not default to fixed anchor-led openers or generic synthesis shells such as "what do the selected materials suggest, show, or imply".
- Use discipline-appropriate biological terms only when they clarify the main object, response, compartment, condition, or assay context; do not add terms just to sound technical.
- Make it clear that the answer depends on the provided materials. You do not need to enumerate every selected figure/table in the stem. If the question remains unambiguous without leading on anchors, do not begin the stem by listing figures/tables. Use specific figure/table labels only when they materially improve disambiguation or local localization.
- Use observation focuses and bridge questions only as latent coverage hints. The stem should still imply an observations -> interpretation -> synthesis dependency, but it does not need to explicitly map or preview each bridge step; it is enough that a respondent can naturally recover that path from the materials. In `bridge_mode = bridge_free`, it must require a local synthesis without inventing a fake intermediate claim.
- Default to figure/table-level anchors. Narrow to panel, timepoint, cell population, or assay slice only when the broader anchor would be misleading or genuinely ambiguous.
- Keep detailed row/column/value checks out of the stem. The stem may be one sentence or multiple sentences; do not force everything into one sentence when that would make the wording denser or harder to parse. Use as many sentences as needed to keep the question natural and readable without turning it into a checklist.

Blind-question constraints:
- Figure/table/panel/location cues and professional terminology are allowed when they help precision, but do not add detail that effectively gives away the target observation.
- Use explicit nouns such as "this interpretation", "this response pattern", "this relationship", or the relevant biological object when a bare pronoun or phrase such as "that account" would be ambiguous.
- If a figure/table is mentioned, use the question-time alias labels supplied by the user JSON. Mention specific labels only when they help precision; otherwise a clear collective reference to the provided materials is acceptable. Do not add unselected anchors or misleading anchor wording.
- Do not directly restate or closely paraphrase final_claim_text or any intermediate claim, and do not expose their key relation structure, directional conclusion, winner shape, or supporting results in advance.
- Do not use proof-task phrasing, exam-essay phrasing, or meta-reasoning protocol language.
- Do not wrap the stem in answer-guiding or meta shells such as "what overall conclusion follows", "what overall picture emerges", "which explanation is best supported", "which interpretation is most consistent with", or "what, if anything, does ...".
- Keep the stem at the level of a real scientific uncertainty, not an instruction about how to solve the QA.
- Mild non-directional guidance is acceptable, but do not reveal result direction, rankings, trend outcomes, intermediate conclusions, or final claim wording.

Return one JSON object only, with no markdown:
{"question":"<blind multihop question>"}

Do not output answers, reasoning, hops, or any extra explanation.

[USER TEMPLATE -- DYNAMIC FIELDS]
article background; selected figure/table aliases; guarded final_claim_text; observation focuses; optional bridge questions; bridge_mode

[OUTPUT CONTRACT]
{"question":"<BLIND_MULTIHOP_QUESTION>"}
Do not output the answer, reasoning steps, involved-reference bookkeeping, or response-requirement bookkeeping.

[DISCIPLINE PROFILE ADAPTATION]
The task and output contract are shared across all four profiles. Only terminology and examples change:
- Biology: biological objects, compartments, conditions, assays, and response patterns.
- Chemistry: compounds, materials, reaction conditions, spectra, morphology, and readouts.
- Computer Science: methods, datasets, metrics, benchmarks, ablations, and failure modes.
- Physics: physical systems, observables, parameters, regimes, spectra, and measurement channels.
\end{PromptCard}
\begin{PromptCard}{C-Q4/5  Question Quality Control and Conditional Rewrite}
[SCOPE]
QC n=234; conditional rewrite n=140

[C-Q5 -- QUESTION QUALITY CONTROL]
Judge whether the blind question is natural, specific, visually grounded, non-leaking, and recoverable from the selected evidence path.

Score only the dimensions used to decide revision:
- naturalness and clarity;
- target specificity;
- anchor localization and selected-material coverage;
- evidence-path recoverability;
- non-leakage and answer-shaping;
- absence of checklist, procedural, or noun-stacked phrasing.

Return JSON containing the dimension scores, failed dimensions, and concise repair advice. Do not answer the scientific question.

[C-Q4 -- CONDITIONAL REWRITE]
Rewrite a blind question into a clearer synthesis question.

You will receive:
- the current question stem;
- focused failure feedback from the rubric and hard checks;
- the selected anchors whose material scope must remain unchanged;
- at most one recent failed candidate summary.

Task: make the smallest useful fix that turns the stem into a clearer, more natural synthesis question while preserving the same anchors, answer contract, and multihop reasoning demand. Do not perform unrelated enhancements.

Requirements:
- Output one rewritten question stem only in JSON form.
- Use the provided failure feedback as the only repair target.
- Do not add new scientific claims, new anchors, or new readout categories that were not needed to fix the feedback.
- Avoid repeating the recent failed candidate phrasing, especially procedural templates, vague takeaway wording, and list-like anchor scaffolds.
- In biology tasks, prioritize these fixes in order when needed:
  1. remove information stacking
  2. remove answer-direction cueing
  3. remove checklist/procedural phrasing
  4. make unnecessarily abstract framing more direct
  5. replace ambiguous pronoun references with explicit nouns
  6. keep only the minimum biological specificity needed to keep the target clear
- Keep the question blind: do not reveal the answer, local outcomes, rankings, or trend directions.
- Keep the selected-material dependency clear, but do not force explicit mention of every selected figure/table. Use specific labels only when they genuinely improve disambiguation or local localization.
- Do not enumerate every evidence unit, row, column, metric, or difference check in the stem.
- Do not add detail that effectively gives away the target observation.
- The rewrite should sound like a peer asking a real biology question after reading the figures, not like a benchmark instruction.
- Preserve a natural evidence-to-interpretation question when it is already clear and non-leaky; rewrite only when the wording sounds templated, compressed, or unnaturally heavy.
- Use at most one or two biological handles that truly help specificity; do not recover specificity by listing every readout family again.
- Prefer a direct key-point question when it makes the stem clearer, but do not collapse it into a vague "What do these figures show?" form.
- Replace unclear references such as "that account" with explicit wording such as "this interpretation", "this response pattern", or the relevant biological object when needed.
- Preserve the need for at least one local interpretive jump before the final synthesis, but do not spell out the reasoning protocol or add extra bridge scaffolding just to make the path more explicit.
- If `bridge_mode` is `bridge_free`, preserve a local synthesis step before the overall answer without introducing a fake intermediate-claim wording.
- Preserve the selected-material scope while changing only the question wording.
- Do not use vague broad anchor phrases that hide selected references.
- Do not directly state or closely paraphrase the hidden final/intermediate claim wording, relation structure, or directional conclusion.
- Do not use exam-essay phrasing such as "How fully is the claim that..." or "to what extent is the claim that...".
- Do not use proof-task phrasing or meta-reasoning protocol phrases such as "using X to form" or "taking X as evidence".
- Do not repair leakage by switching into shells such as "what overall conclusion follows", "which explanation is best supported", "which interpretation is most consistent with", or "what, if anything, does ...".
- Do not fix naturalness by merely shortening the stem into something too vague; shorter is not automatically better.
- If the current stem compresses too much information into one sentence, it is acceptable to split it into multiple sentences when that improves readability and still sounds like one natural research question.

Return one JSON object only, with no markdown:
{"question":"<rewritten blind multihop question>"}

[INPUT / OUTPUT]
Input: current question, failed QC dimensions, repair advice, selected anchors, and the unchanged answer contract.
Output: {"question":"<MINIMALLY_REWRITTEN_BLIND_QUESTION>"}

[DISCIPLINE PROFILE ADAPTATION]
The task and output contract are shared across all four profiles. Only terminology and examples change:
- Biology: biological objects, compartments, conditions, assays, and response patterns.
- Chemistry: compounds, materials, reaction conditions, spectra, morphology, and readouts.
- Computer Science: methods, datasets, metrics, benchmarks, ablations, and failure modes.
- Physics: physical systems, observables, parameters, regimes, spectra, and measurement channels.
\end{PromptCard}
\begin{PromptCard}{C-Q6  Gold-Answer Generation}
[SCOPE]
stage observed in n=235 benchmark samples

[SYSTEM PROMPT]
You are writing the standard answer for an academic reading assessment.

You will receive:
- the question stem;
- the program-extracted preferred claim;
- the program-provided canonical trajectory, which is the gold dependency chain and must not be altered in node identity or order.

Task: use the canonical trajectory and execution plan to write one clear, checkable standard answer.

Requirements:
- The final answer must resolve to the claim identified by `preferred_claim_id`.
- For every step, clearly state which figure/table ids and established bases it depends on, and what evidence or claim it derives.

Source-fidelity requirements:
- Follow every evidence-to-claim dependency in the supplied canonical trajectory internally before writing the answer.
- Ground evidence and claim wording in their `article_source_attributions`; use each exact excerpt only as verification context and do not copy irrelevant surrounding text.
- Attribute figure/table observations to the corresponding task aliases and make the final evidence-to-claim connection explicit.

Return one JSON object only, with no markdown:
{"answer":"<complete gold-standard answer>"}

Bad examples:
- Rewriting or replacing `derived_node_id` values from the canonical trajectory.

[USER TEMPLATE -- DYNAMIC FIELDS]
question; selected figure/table context; canonical evidence-to-claim steps; established-basis context; node_source_attributions for evidence and claims

[OUTPUT CONTRACT]
{"standard_answer":"<GOLD_ANSWER_GROUNDED_IN_THE_CANONICAL_PATH>"}
Use article-source excerpts to verify scientific meaning, but write the answer for a respondent who sees the selected visual materials rather than the source article text.
\end{PromptCard}
\subsection{Reported Inference Settings}

\begin{PromptCard}{R1  LLM Reasoning}
[SCOPE]
eight models x 235 samples; audited records n=1880

[SYSTEM PROMPT]
You are a careful scientific figure QA model. Answer concisely.

[USER TEMPLATE]
Answer the question using the provided figures/tables.

Question:
<QUESTION>

[MULTIMODAL CONTENT PARTS]
Figure 1:
<FIGURE_1_PIXELS>
Table 1:
<TABLE_1_PIXELS>
\end{PromptCard}
\begin{PromptCard}{R2  Evidence Hint}
[SCOPE]
eight models x 235 samples; audited records n=1880

[SYSTEM PROMPT]
You are a careful scientific figure QA model. Answer concisely.

[USER TEMPLATE]
Answer the question using the provided figures/tables.

Question:
<QUESTION>

Evidence hints:
- Evidence <STEP_ID> (<SOURCE_REFERENCE_LABELS>): <DERIVED_EVIDENCE_STATEMENT>
- <ADDITIONAL_EVIDENCE_HINTS_AS_NEEDED>

Use the evidence hints to focus on relevant visual evidence, but verify them against the provided figures/tables before answering.

[MULTIMODAL CONTENT PARTS]
Figure 1:
<FIGURE_1_PIXELS>
Table 1:
<TABLE_1_PIXELS>
\end{PromptCard}
\begin{PromptCard}{R3  Agentic Tool-Use}
[SCOPE]
canonical protocol plus serialization/continuation variants

[CANONICAL FUNCTION-TOOL PROTOCOL]
You are a crop-only scientific figure/table QA agent.

You must answer using the provided figure/table pixels and crop operations only.
On every turn, call exactly one available tool. Every tool call must include `reasoning` explaining your thinking.

Image visibility:
- Images are attached directly in each model turn, alongside the Available images list.
- Original figure/table images are visible from the first turn.
- crop_image creates a new listed crop image; that crop is attached directly in later turns.
- Keep key visual facts in reasoning and, when available, image captions so later steps stay grounded.

Available actions:

1. crop_image
Crop a specific region of a provided figure/table.
Use it only when closer inspection of labels, values, legends, axes, panels, or table cells could affect the answer.
Crop only original figure/table images, not crop outputs such as figure1_crop1.
The bbox must be normalized [x1, y1, x2, y2] with values in [0, 1].
Coordinates use the source image's top-left origin: (0, 0) is the top-left, x increases to the right, and y increases downward.
Do not use crop_image for references that are not listed in the task.
Schema:
{"action_type":"crop_image","reasoning":"inspect the answer-relevant legend and axis labels more closely","image_name":"figure1","bbox":[0.3,0.4,0.5,0.6]}

2. finish
Use this when the current visual evidence and crop observations are sufficient to answer the user question, and additional crops are unlikely to change the core conclusion or add an important qualification.
The final answer should explain the key figure/table observations first, then any external context that materially changes or qualifies the interpretation, then reason through intermediate interpretations step by step before giving the synthesized conclusion. When relevant, also include what new knowledge the exploration uncovered beyond the literal question, plus uncertainties, alternative explanations, worthwhile follow-up questions, possible experimental concerns, or future research directions suggested by the figure/table.
Do not mention hidden gold answers or internal task fields.
Schema:
{"action_type":"finish","reasoning":"explain why the available visual evidence is sufficient to answer","text":"final answer..."}

Decision policy:
- Images are attached directly in every turn; inspect the attached figure/table pixels before choosing an action.
- Use crop_image when closer visual/table inspection is needed.
- Do not use search, fetch_page, fetch_image, or update_image_caption; they are unavailable in this profile.
- Do not rely on external web evidence, page text, or hidden metadata.
- Avoid repeated crops of the same area unless the previous crop was unreadable.
- Continue cropping only if the next crop could change the core conclusion or add an important qualification.
- Finish once the remaining uncertainty would only add minor detail or confidence.

Output format:
Call exactly one tool matching one of the schemas above.

[AVAILABLE TOOLS]
crop_image(image_name, bbox, reasoning): inspect an answer-relevant region using normalized [x1,y1,x2,y2] coordinates.
finish(text, reasoning): submit the final answer when further crops are unlikely to change or materially qualify it.

[USER TEMPLATE]
Question: <QUESTION>
Available images: <ORIGINAL_FIGURE_AND_TABLE_NAMES>
Inline pixels: <ORIGINAL_IMAGES_AND_ONE_TURN_CROPS>

[RECORDED PROTOCOL VARIANTS]
The reported trajectories include the dominant function-tool protocol, a small recovered-shard function variant, and a ReAct-JSON serialization variant. They expose the same crop-only evidence-access policy. Conditional continuation messages only enforce remaining-turn and invalid-action constraints; they do not add scientific evidence.
\end{PromptCard}
\subsection{Evaluation}

\begin{PromptCard}{E1  Evidence and Intermediate-Claim Coverage}
[SCOPE]
produces E-Cov. and C-Cov.; source: figure_qa_eval/core/graders.py

[SYSTEM PROMPT]
You are a scientific QA graph-structure judge. Your task is NOT to score overall answer quality. You only judge whether the agent's answer substantively covers the specified evidence nodes and intermediate claims.

RULES:
1. You are judging structural coverage, not overall quality or correctness.
2. Allow semantic equivalence -- the answer does not need to repeat node labels verbatim. If the answer's content supports the core scientific meaning of a node, mark it as matched.
3. A node is matched only when the answer substantively engages with its scientific point. Vague or tangential mentions do not count.
4. Only output node IDs that appear in the input. Never invent new node IDs.
5. Do NOT output coverage scores -- only node-level yes/no judgments.
6. Treat article source attributions only as provenance context for interpreting the intended node meaning. The answer need not quote an excerpt or name its section to receive a yes judgment.

COMPACT OUTPUT RULES:
1. Output exactly one JSON object and nothing else.
2. Consider the full input carefully before assigning labels.
3. Do not restate or quote the question, answer, hints, evidence, or input.
4. Use only the keys shown in the schema example.
5. Each judgment entry should contain only the schema-required keys.

OUTPUT FORMAT: You must output exactly one JSON object with this schema:
{"evidence_node_judgments": [{"node_id": "id1", "raw_label": "yes"}, ...], "intermediate_claim_judgments": [{"node_id": "id3", "raw_label": "yes"}, ...]}

All node IDs must only come from the input. raw_label must be "yes" or "no". Output raw JSON without markdown code fences.

[USER TEMPLATE]
{
  "question": "<QUESTION>",
  "answer_text": "<MODEL_ANSWER>",
  "required_evidence_nodes": [
    {
      "node_id": "<EVIDENCE_NODE_ID>",
      "label": "<EVIDENCE_NODE_LABEL>",
      "citation": "<CITATION_OR_NULL>",
      "original_label": "<ORIGINAL_EVIDENCE_LABEL>",
      "article_source_attributions": [
        {
          "section_title": "<SOURCE_SECTION_OR_NULL>",
          "source_excerpt": "<SHORT_VERBATIM_EVIDENCE_PASSAGE>"
        }
      ]
    }
  ],
  "required_intermediate_claims": [
    {
      "node_id": "<INTERMEDIATE_CLAIM_NODE_ID>",
      "label": "<INTERMEDIATE_CLAIM_LABEL>",
      "article_source_attributions": [
        {
          "section_title": "<SOURCE_SECTION_OR_NULL>",
          "source_excerpt": "<SHORT_VERBATIM_CLAIM_PASSAGE>"
        }
      ]
    }
  ]
}
\end{PromptCard}
\begin{PromptCard}{E2  Final-Answer Correctness}
[SCOPE]
produces Accuracy; source: figure_qa_eval/core/graders.py

[SYSTEM PROMPT]
You are a scientific QA correctness judge. Your ONLY task is to judge whether the agent's final answer is semantically correct compared to the gold standard answer.

CRITICAL: The agent does NOT need to explicitly state evidence, intermediate claims, or reasoning steps. As long as the final conclusion is semantically correct, the answer is correct.
The gold standard answer is the canonical reference and may be more specific than a compressed final claim, so compare against what the standard answer actually says.

Judge these sub-items (binary: yes/no/null):

1. gives_final_conclusion_to_question (binary): Does the answer explicitly give a final conclusion that addresses the main question?
2. matches_gold_final_claim (binary, legacy key name): Is the final conclusion semantically consistent with the gold standard answer?
3. preserves_direction_polarity_comparison (binary): When the gold standard answer contains direction/polarity/comparison, does the answer preserve it? Use null if the gold standard answer does not involve direction/polarity/comparison.
4. does_not_make_materially_conflicting_conclusion (binary): Does the answer NOT make any conclusion that materially conflicts with the gold standard answer?

Scoring: binary: "yes"=1.0, "no"=0.0, "null"=not applicable.

COMPACT OUTPUT RULES:
1. Output exactly one JSON object and nothing else.
2. Consider the full input carefully before assigning labels.
3. Do not restate or quote the question, answer, hints, evidence, or input.
4. Use only the keys shown in the schema example.
5. Each judgment entry should contain only the schema-required keys.

OUTPUT FORMAT:
{"item_scores": {"gives_final_conclusion_to_question": {"raw_label": "yes"}, "matches_gold_final_claim": {"raw_label": "yes"}, "preserves_direction_polarity_comparison": {"raw_label": "yes"}, "does_not_make_materially_conflicting_conclusion": {"raw_label": "yes"}}}

Output raw JSON without markdown code fences.

[USER TEMPLATE]
{
  "question": "<QUESTION>",
  "gold_standard_answer": "<GOLD_STANDARD_ANSWER>",
  "answer_text": "<MODEL_ANSWER>"
}
\end{PromptCard}
\flushbottom